\documentclass{article} % For LaTeX2e
\usepackage{iclr2025_conference,times}

\usepackage{amsmath,amsfonts,bm}

\def\eqref#1{equation~\ref{#1}}
\def\1{\bm{1}}

\DeclareMathAlphabet{\mathsfit}{\encodingdefault}{\sfdefault}{m}{sl}
\SetMathAlphabet{\mathsfit}{bold}{\encodingdefault}{\sfdefault}{bx}{n}

\usepackage{hyperref}
\usepackage{url}
\usepackage{subfigure}
\usepackage{makecell}
\usepackage{wrapfig,lipsum}
\usepackage{bbold}
\usepackage{url}
\usepackage{hyperref}
\usepackage{CJKutf8}
\usepackage{graphicx}
\usepackage{times}
\usepackage{pdfpages}
\usepackage{amsmath}
\usepackage{tablefootnote}
\usepackage{tcolorbox}
\usepackage{colortbl}
\usepackage{float}
\usepackage{multirow}
\usepackage{booktabs}
\usepackage{latexsym}
\usepackage{listings}
\usepackage{bbding}
\usepackage{pifont}
\usepackage{wasysym}
\usepackage{amssymb}
\usepackage{paralist}
\usepackage{spverbatim}
\usepackage{fancyvrb}
\usepackage{fvextra}

\usepackage{caption}
\usepackage{subcaption}

\usepackage{tikz}
\usepackage{longtable}
\renewcommand{\footnotesize}{\fontsize{9pt}{11pt}\selectfont}

\colorlet{shadegray}{gray!20}
\colorlet{darkgray}{gray!50}
\definecolor{cloud}{HTML}{C4C6D0}
\definecolor{silver}{HTML}{9090C0}
\definecolor{dolphin}{HTML}{5C5858}
\definecolor{lightgray}{HTML}{D3D3D3}
\definecolor{mediumgray}{HTML}{BEBEBE}
\definecolor{sonic}{HTML}{757575}
\usepackage{enumitem}

\title{Training Language Models to Critique With Multi-agent Feedback}

\author{Tian Lan$^{1*}$ \quad Wenwei Zhang$^{2}$\thanks{\quad Equal contributions} \quad Chengqi Lyu \quad Shuaibin Li \quad Chen Xu$^{3}$ \quad Heyan Huang$^{1}$ \\
\textbf{Dahua Lin$^{2}$ \quad Xian-Ling Mao$^{1}$\thanks{\quad Corresponding author} \quad Kai Chen$^{2 \dag}$}\\
$^{1}$School of Computer Science and Technology, Beijing Institute of Technology\\
$^{2}$Shanghai AI Laboratory\\
$^{3}$School of Medical Technology, Beijing Institute of Technology\\
\texttt{lantiangmftby@gmail.com}
}

\newcommand\redcolor[1]{\cellcolor{gray!30!red!30}{#1}}
\newcommand\bluecolor[1]{\cellcolor{gray!30!blue!30}{#1}}

\iclrfinalcopy % Uncomment for camera-ready version, but NOT for submission.
\begin{document}

\maketitle

\begin{abstract}
Critique ability, a meta-cognitive capability of humans, presents significant challenges for LLMs to improve.
Recent works primarily rely on supervised fine-tuning (SFT) using critiques generated by a single LLM like GPT-4.
However, these model-generated critiques often exhibit flaws due to the inherent complexity of the critique. 
Consequently, fine-tuning LLMs on such flawed critiques typically limits the model's performance and propagates these flaws into the learned model.
To overcome these challenges, this paper proposes a novel data generation pipeline, named MultiCritique, that improves the critique ability of LLMs by utilizing multi-agent feedback in both the SFT and reinforcement learning (RL) stages.
First, our data generation pipeline aggregates high-quality critiques from multiple agents instead of a single model, with crucial information as input for simplifying the critique.
Furthermore, our pipeline improves the preference accuracy of critique quality through multi-agent feedback, facilitating the effectiveness of RL in improving the critique ability of LLMs.
Based on our proposed MultiCritique data generation pipeline, we construct the MultiCritiqueDataset for the SFT and RL fine-tuning stages. Extensive experimental results on two benchmarks demonstrate:
1) the superior quality of our constructed SFT dataset compared to existing critique datasets;
2) additional improvements to the critique ability of LLMs brought by the RL stage.
Notably, our fine-tuned 7B model significantly surpasses other advanced 7B-13B open-source models, approaching the performance of advanced 70B LLMs and GPT-4.
Codes, datasets and model weights will be publicly available.
\end{abstract}
% !TEX root = ../main.tex
\section{Introduction}

The critique ability, \textit{i.e.}, the capability to identify and refine flaws in responses, has been widely used to facilitate reliable automatic evaluation and self-improvement of LLMs~\citep{lan2024criticbenchevaluatinglargelanguage,wu2024metarewardinglanguagemodelsselfimproving}.
As a meta-cognitive capability~\citep{toy2024metacognition,wang2024metacognitivepromptingimprovesunderstanding}, critique ability requires LLMs to possess a deep understanding of user queries and evaluated responses beyond mere criticism~\citep{kim2024prometheusinducingfinegrainedevaluation,zheng2023judgingllmasajudgemtbenchchatbot}. 
Therefore, it is challenging to improve the critique ability of LLMs~\citep{lan2024criticbenchevaluatinglargelanguage,lin2024criticbenchbenchmarkingllmscritiquecorrect}.

Recent researches indicate that open-source LLMs usually demonstrate weaker critique abilities compared to advanced closed-source models~\citep{lan2024criticbenchevaluatinglargelanguage,lin2024criticbenchbenchmarkingllmscritiquecorrect}.
Therefore, numerous works have been proposed to improve the critique ability of LLMs, primarily through Supervised Fine-Tuning (SFT) using critiques generated by one single LLM, typically GPT-4~\citep{li2024generative,cui2023ultrafeedback}.
However, these model-generated critiques often exhibit flaws like inaccurate descriptions or suggestions about flaws in responses due to the complexity of the critique~\citep{verga2024replacingjudgesjuriesevaluating,lan2024criticbenchevaluatinglargelanguage,liu2024empiricalanalysislargelanguage}.
Consequently, the fine-tuned LLMs on these datasets are constrained by these flawed critiques, and these flaws in critiques are propagated into the learned models during the SFT stage, leading to the potential amplification of these issues.

\begin{figure*}[t]
\center{\includegraphics[width=0.98\textwidth]{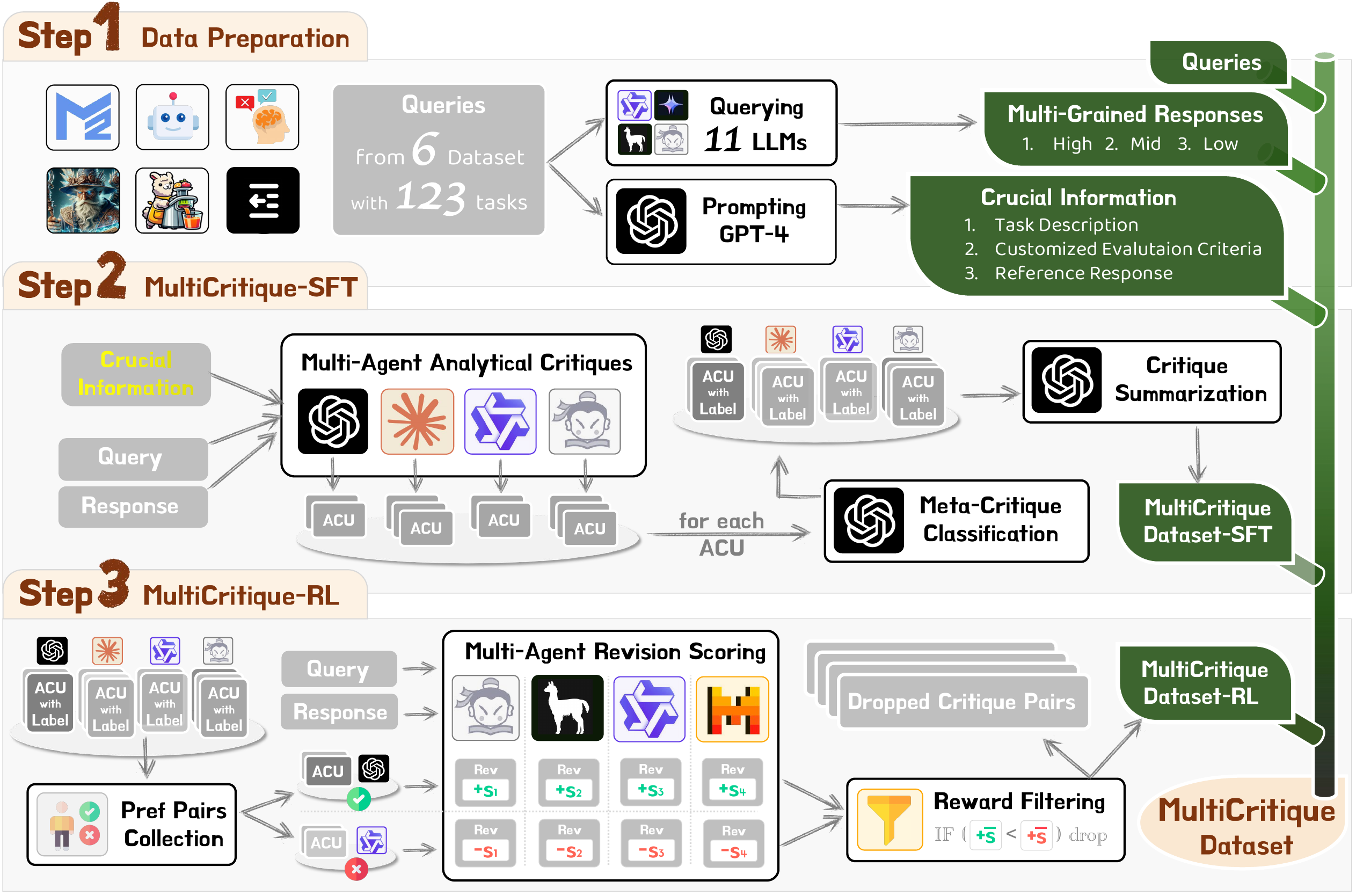}}
\vspace{-6pt}
    \caption{The overview of our proposed MultiCritique data generation pipeline. First, we prepare queries and evaluate responses and crucial information (Step 1). Then, we conduct the MultiCritique-SFT pipeline to construct the high-quality SFT critique dataset (Step 2). Finally, we conduct the MultiCritique-RL pipeline to construct the preference critique dataset for the RL stage (Step 3). An ACU is a structured unit for identifying one specific flaw in the evaluated response. A list of model-generated ACUs denotes the analytical critique.}\label{img:overview}
\vspace{-12pt}
\end{figure*}

To address these challenges, in this paper, we develop a novel data generation pipeline named \textbf{MultiCritique}. This approach utilizes multi-agent feedback to construct the high-quality dataset for the SFT and RL fine-tuning stages.
First of all, to mitigate the effects of flawed critiques generated by one single LLM, we develop the \textbf{MultiCritique-SFT} pipeline, as shown in Figure~\ref{img:overview} (Step 2). 
Specifically, it first collects the multi-agent analytical critiques from four advanced LLMs rather than one single LLM, with three crucial information as input for simplifying the critique: (1) A detailed task description of the user query; (2) A customized two-tier structured evaluation criteria; and (3) A reference response tailored to satisfy these criteria.
Each model-generated analytical critique consists of a list of \textbf{A}nalytical \textbf{C}ritique \textbf{U}nits (ACUs), with each ACU identifying one specific flaw in the evaluated responses.
Then, GPT-4 conducts the meta-critique classification process to classify these ACUs into seven quality categories automatically~\citep{lan2024criticbenchevaluatinglargelanguage}, given all multi-agent analytical critiques as context.
The process concludes by summarizing a final analytical critique that aggregates high-quality ACUs while discarding flawed ones, effectively combining the advantages of multiple models.

Second, to move beyond the mere behavior cloning on model-generated critiques in the SFT dataset, 
we develop the \textbf{MultiCritique-RL} data generation pipeline to construct the high-quality preference critique dataset via multi-agent feedback, facilitating the effectiveness of RL in improving the critique ability of LLMs.
Specifically, as shown in Figure~\ref{img:overview} (Step 3), chosen (\textbf{+}) and rejected (\textbf{-}) analytical critiques are naturally paired via meta-critique classification in the MultiCritique-SFT pipeline, where chosen analytical critiques exhibit fewer and minor flawed ACUs than rejected ones.
However, the preference between these chosen and rejected analytical critiques might be inaccurate due to the complex meta-critique analysis and limited model capability~\citep{lan2024criticbenchevaluatinglargelanguage,sun2024critiquecritique}.
To address this issue, we then propose the \textbf{M}ulti-\textbf{A}gent-\textbf{R}evision-\textbf{S}coring (MARS) filtering to refine the preference critique dataset, ensuring the chosen analytical critiques lead to superior revisions compared to rejected ones across multiple models.
These preference pairs are used to fine-tune a reward model, guiding the RL process to improve critique ability further.
Unlike recent works that utilize preference learning to improve the critique ability of LLMs, such as Themis~\citep{hu2024themis} and CriticGPT~\citep{mcaleese2024llmcriticshelpcatch}, our proposed MultiCritique-RL pipeline does not require any human annotations, thereby demonstrating better scalability.

Based on our proposed MultiCritique data generation pipeline, we construct the \textbf{MultiCritiqueDataset} for improving critique ability, consisting of two splits for the SFT and RL fine-tuning stages (right part of Figure~\ref{img:overview}).
Extensive experiments on two benchmarks demonstrate the superior quality of our proposed MultiCritiqueDataset compared to other critique datasets. 
Specifically, even if the scales of existing datasets are over three and eight times larger than our proposed dataset, the model fine-tuned during the SFT stage on our proposed MultiCritiqueDataset still significantly outperforms those fine-tuned on other datasets, with 21.48\% and 22.50\% average performance gain on \textsc{CriticEval}~\citep{lan2024criticbenchevaluatinglargelanguage} and \textsc{CriticBench}~\citep{lin2024criticbenchbenchmarkingllmscritiquecorrect} benchmarks, respectively. 
Furthermore, the RL fine-tuning stage on MultiCritiqueDataset further boosts the critique ability of the SFT model, resulting in 6.3\% and 0.51\% absolute improvements on \textsc{CriticEval} and \textsc{CriticBench}.
Besides, ablation studies confirm the positive contributions of crucial information in simplifying the critique and the roles of our proposed MultiCritique data generation pipeline in enhancing dataset quality.
In summary, the InternLM2-7B-Chat-SFT model that fine-tuned by SFT and RL stages on our proposed MultiCritiqueDataset, significantly outperforms advanced 7B-13B instruction-tuned and critique-tuned LLMs as well as the GPT-3.5-Turbo model, approaching advanced 70B LLMs and GPT-4. For example, our model approaches GPT-4 on \textsc{CriticBench}, with an F1 score of 75.66\% vs. 78.75\%, significantly outperforming average performance of 7B-13B instruction-tuned and critique-tuned LLMs (55.58\%). 
% !TEX root = ../main.tex
\section{Related Work}

\noindent\textbf{Critique Ability of LLMs}
The critique ability of LLMs has been applied in three key areas:
(1) Reliable Automatic Evaluation: LLMs can achieve high correlation with human annotators in response evaluation~\citep{liu2023gevalnlgevaluationusing,zheng2023judgingllmasajudgemtbenchchatbot};
(2) Self-Improvement of LLMs: This ability facilitates the self-improvement of LLMs during inference and training stages~\citep{yuan2024selfrewardinglanguagemodels,wu2024metarewardinglanguagemodelsselfimproving};
(3) Robust Reward Modeling: Textual critiques contribute to robust reward modeling by providing detailed feedback~\citep{ye2024improvingrewardmodelssynthetic,zhang2024generativeverifiersrewardmodeling}.
Recently, to bridge the gap between the weak open-source LLMs and advanced closed-source models,
various datasets have been developed to enhance the critique capability of LLMs by mimicking GPT-4 behavior~\citep{lan2024criticbenchevaluatinglargelanguage,lin2024criticbenchbenchmarkingllmscritiquecorrect}.
For example, datasets such as UltraFeedback~\citep{cui2023ultrafeedback}, Auto-J~\citep{li2024generative} and Prometheus~\citep{kim2024prometheusinducingfinegrainedevaluation}, collect critiques from GPT-4 to fine-tune open-source LLMs like Llama2-13B.
Despite their potential, these datasets often suffer from quality issues due to the inherent complexity of critiques~\citep{verga2024replacingjudgesjuriesevaluating}.
Compared with these existing works, we first propose a novel data generation pipeline, MultiCritique, to collect high-quality critiques from multi-agents.
Then, reinforcement learning is used to improve the critique ability of LLMs, moving beyond the mere behavior cloning on the supervised critique dataset.

%\noindent\textbf{Preference-based Reinforcement Learning (RL)}
\noindent\textbf{Preference-based Reinforcement Learning (RL)}
Reinforcement learning (RL) algorithms, like PPO~\citep{schulman2017proximalpolicyoptimizationalgorithms}, are widely utilized to guide LLMs to generate responses that are more preferred by humans~\citep{yang2024rlcdreinforcementlearningcontrastive,xu2024chatglmmathimprovingmathproblemsolving,yuan2024selfrewardinglanguagemodels}. 
RL algorithms typically employ a reward model as a proxy for human judgment, learning through human-annotated pairwise comparison of responses. This is often called Reinforcement Learning from Human Feedback (RLHF)~\citep{stiennon2022learningsummarizehumanfeedback}. 
Recently, some works have been proposed to improve the critique ability through reinforcement learning. For example, CriticGPT~\citep{mcaleese2024llmcriticshelpcatch} improves the critique ability of ChatGPT through RLHF. 
Similarly, Themis~\citep{hu2024themis}, Self-Taught Evaluator~\citep{wang2024selftaughtevaluators}, SFR~\citep{wang2024directjudgementpreferenceoptimization} collect the preference critique pairs by examining consistency between the LLM's judgment scores and the human-annotated judgment scores.
However, their solution is limited due to the high cost of human annotations. In contrast, our proposed MultiCritique pipeline automatically collects preference critique datasets via a multi-agent framework without human supervision.
Our concurrent work, Meta-Rewarding~\citep{wu2024metarewardinglanguagemodelsselfimproving}, collects preference critique pairs using optimized LLM as the meta-judge model.
Compared with Meta-Rewarding, we propose the Multi-Agent-Revision-Scoring (MARS) to filter high-quality critique pairs by using the quality of their revisions as an indicator of critique quality, thereby facilitating a more robust RL.

%\noindent\textbf{Multi-agent Framework}
\noindent\textbf{Multi-agent Framework}
The multi-agent framework has been shown to foster more divergent thinking and aggregate the diverse opinions of multiple LLMs~\citep{liang2024encouragingdivergentthinkinglarge,zhang2024exploringcollaborationmechanismsllm}, effectively improve the alignment of LLMs. For example, Arena learning~\citep{luo2024arenalearningbuilddata} and Stable Alignment~\citep{liu2024training} improve the alignment of LLMs via multi-agent battle results or social simulation.
Recently, the multi-agent framework has also been used for improving critique ability. For exmaple, PoLL~\citep{verga2024replacingjudgesjuriesevaluating} addresses the model-specific biases by pooling judgment scores of a panel of small models.
ChatEval~\citep{chan2023chateval} enhances the reliability of LLM-based automatic evaluations via multi-agent debate. 
However, the quality of their multi-agent critiques are not assessed, which may lead to incorrect critiques influencing other LLMs.
Compared to their work, we explicitly evaluate the quality of critiques through (1) GPT-4-based meta-critique; and (2) evaluating revision quality as an indicator of critique quality.
% !TEX root = ../main.tex

\section{Method}
\subsection{Data Preparation}

%In this section, we will introduce the data generation pipeline for constructing the \textsc{MultiCritique} datasets for the SFT and RL phases. It involves four sequential steps: (1) \textbf{Data Preparation} gathers diverse queries and evaluated responses; (2) \textbf{Crucial Information Collection} elicit crucial information to simplify critique task; (3) \textbf{Multi-Agent-Critique-Analyze-Summarize (MACAS) Pipeline} collects critiques from diverse multi-agent perspectives, and consolidate them into a final critique for the SFT phase, given meta-critique analyses; (4) \textbf{Preference Dataset Construction} finally construct preference critique pairs for RL phase. The overview of construction is shown in Figure~\ref{img:overview}.

As shown in Figure~\ref{img:overview} (Step 1), we collect diverse queries and evaluated responses before running our proposed MultiCritique data generation pipeline. Besides, we also elicit crucial information from the queries to simplify the critique. Details are elaborated as follows.

%% 可以根据appendix内容扩充
\noindent\textbf{Diverse Queries Collection} We first compile queries from several well-established datasets. These include alignment datasets such as OpenHermes-2.5, DEITA~\citep{liu2024what} and OpenAssistant~\citep{köpf2023openassistantconversationsdemocratizing};
mathematical and coding reasoning datasets like MetaMathQA~\citep{yu2024metamathbootstrapmathematicalquestions} and CodeFeedback~\citep{zheng-etal-2024-opencodeinterpreter};
as well as the critique dataset Auto-J~\citep{li2024generative}.
In total, we sample 10.7K queries covering 123 diverse task scenarios.

\noindent\textbf{Diverse Evaluated Responses Collection} 
Once the queries are collected, we use eleven LLMs with different capabilities to generate responses with varying response qualities. These responses are then evaluated using the robust reward model\textemdash InternLM2-20B-reward~\citep{cai2024internlm2}, which performs well in RewardBench~\citep{lambert2024rewardbenchevaluatingrewardmodels}.
We select three low-, medium- and high-quality responses for each query, ensuring a significant performance gap.
In total, 10.7K$\times$3=32.1K query-response pairs are collected.

\noindent\textbf{Crucial Information Collection}
\label{sec:collect_crucial_information}
Once the query-response pairs are collected, we elicit three crucial information sequentially using GPT-4 to simplify the critique:
\textbf{(1) Task Description}: Our preliminary findings indicate that LLMs may misunderstand the objectives of queries. By prompting GPT-4 to describe the task, we can mitigate this issue to some extent;
\textbf{(2) Customized Evaluation Criteria}:
Once the task description is obtained, we propose generating customized two-tier structure evaluation criteria tailored to each query to guide effective critiques.
The first tier outlines the fundamental evaluation dimensions for the task, while the second tier offers more customized criteria. Each criterion is structured with a name, description and level of importance~\citep{li2024generative};
\textbf{(3) Reference Response}: Finally, we generate reference responses that satisfy all customized evaluation criteria. Since reference responses tend to produce critiques that lack diversity for mathematical and coding questions, we set reference answers as empty for these two tasks.

%Differences from previous works in crucial information are detailed in Appendix~\ref{sec:difference_between_recent_works}.
Following these three steps, we collect $\mathbb{N}$=32.1K samples $\{(q_i,r_i,\mathcal{CI}_i)\}_{i=1}^{\mathbb{N}}$, where $q_i,r_i,\mathcal{CI}_i$ represents the $i$-th query, evaluated response and corresponding crucial information in the dataset. Please refer to Appendix~\ref{appendix:implementation_details} for more details and statistical information about the collection of queries and evaluated responses.

\subsection{MultiCritique-SFT Data Generation Pipeline}
After collecting query-response and crucial information, we introduce our proposed MultiCritique-SFT data generation pipeline to construct the high-quality SFT dataset. 
The overview of MultiCritique-SFT pipeline is shown in Figure~\ref{img:overview} (Step 2).
This pipeline gathers multi-agent critiques to mitigate flawed critiques generated by one single LLM.
%the model-specific bias during critique.
%It consists of three sequential steps: (1) Multi-agent Analytical Critique; (2) Meta-critique Analyze and (3) Critique Summarize. In this pipeline, multiple LLMs first provide detailed analytical critiques simultaneously. Subsequently, GPT-4 conduct the meta-critique to assess these critiques. The final critique is then summarized by considering all the multi-agent critiques and their corresponding meta-critique analyses.

To achieve this goal, we aim to aggregate the high-quality content in multi-agent critiques into a comprehensive and accurate critique.
Therefore, we first collect detailed analytical critiques simultaneously from multiple LLMs (\textbf{Multi-Agent Analytical Critique}). Each analytical critiques consist of a list of \textbf{A}nalytical \textbf{C}ritique \textbf{U}nits (ACUs). Then, the GPT-4 classifies each ACU into one of seven quality categories (\textbf{Meta-Critique Classification}).
Finally, a final critique is aggregated by summarizing the accurate ACU while discarding the flawed ones (\textbf{Critique Summarization}).
Unlike previous works allowing multi-agent discussions or debates~\citep{chan2023chateval}, our preliminary study observes that LLM's critiques could be easily influenced by other critiques, potentially reducing the diversity and comprehensiveness of critiques.  
Thus, we do not allow the inner-model discussions. Instead, we utilize GPT-4 as a summarizer to evaluate each critique content given all LLMs critiques as context, since previous works~\citep{lan2024criticbenchevaluatinglargelanguage,lin2024criticbenchbenchmarkingllmscritiquecorrect} have proven that only GPT-4 could effectively conduct the meta-critique.

\noindent\textbf{Multi-agent Analytical Critiques}
\label{sec:multi_agent_analytical_critique}
We employ four LLMs to simultaneously critique the evaluated responses: GPT-4, Claude-1-instant, Qwen-1.5-72B-Chat and InternLM2-20B-Chat, which exhibits strong performance on \textsc{CriticEval} benchmark~\citep{lan2024criticbenchevaluatinglargelanguage}. 
Each LLM performs sentence-by-sentence and cross-sentence critique and generates a structured analytical critique, which is a list of \textbf{A}nalytical \textbf{C}ritique \textbf{U}nits (ACUs).
An ACU is a structured unit for identifying and addressing a specific flaw in the evaluated response, consisting of five values about the flaw: 
(1) the location\footnote{We introduce a pre-processing step to label sentences in evaluated responses, as detailed in  Appendix~\ref{appendix:response_pre_process}.}; 
(2) the description; 
(3) suggestions for revision; 
(4) the criteria type; 
and (5) the severity.
The structured ACUs not only benefits the robust subsequent meta-critique process~\citep{sun2024critiquecritique} but also exhibit great interprebability.
%Analytical critiques of these four models are obtained, and each encompasses a list of ACUs for pointing out flaws in the evaluated responses.

\iffalse
Overall, the critique $C_{i,j}$ of $j$-th LLM for $i$-th sample is obtained, consisting of a list of ACUs: 
\begin{equation}
    C_{i,j}=\{{\text{ACU}}_{i,j,k}\}^{N_{i,j}}_{k=1}=\boldsymbol{{\text{LLM}}_{j}}(\mathcal{P}_{\text{cri.}}, q_i, r_i, \mathcal{CI}_i)
\end{equation}
where $N_{i,j}$ denotes the number of ACUs. $\mathcal{P}_{\text{cri.}}$ is the designed prompt detailed in Appendix~\ref{appendix:prompt_critique}.
\fi

\noindent\textbf{Meta-Critique Classification}
\label{sec:meta_critique_labeling}
Since LLMs often produce flawed critiques due to the limited model capability~\citep{lan2024criticbenchevaluatinglargelanguage,wang2023shepherd}, it is crucial to identify these flaws before aggregating them into a final comprehensive and accurate analytical critique.
To achieve this goal, we employ GPT-4 to judge the quality of these model-generated analytical critiques, a concept known as meta-critique~\citep{sun2024critiquecritique,lan2024criticbenchevaluatinglargelanguage}.
Specifically, GPT-4 classifies each ACU into one of seven quality categories by considering all the multi-agent analytical critiques in the context.
These quality categories are determined by human annotators and are associated with human-annotated severity scores ranging from 1 to 5. 
Each quality category reflects one specific flaw in an ACU.
Please refer to Appendix~\ref{appendix:prompt_meta_critique} for more details.
Importantly, for one model-generated analytical critique, the accumulated severity scores of its ACUs could indicate its quality, whereas a higher accumulated severity score indicates lower analytical critique quality.

\noindent\textbf{Critique Summarization}\label{sec:critique_summarize}
After the previous steps, GPT-4 summarizes these ACUs into a comprehensive analytical critique.
This process involves retaining and merging accurate ACUs generated by multiple LLMs while modifying or excluding those identified as flawed.
Except for the final summarized analytical critique, we also prompt GPT-4 to generate an overall description and judgment score for the evaluated response.
Finally, the final analytical critique, description and judgment score are concatenated as the final critique for the evaluated response, denoted as $\mathcal{C}$.

By following previous steps of MultiCritique-SFT, we construct a supervised fine-tuning dataset MultiCritiqueDataset-SFT, consisting of $\mathbb{N}$=32.1K samples: $\{(q_i,r_i,\mathcal{CI}_i,\mathcal{C}_i)\}_{i=1}^{\mathbb{N}}$.

\iffalse
Overall, the meta-critique analysis for the ACU list in $j$-th model critique of $i$-th sample is represented as:
\begin{equation}
    {\text{Meta}}\text{-}C_{i,j}=\{(\boldsymbol{l}_{i,j,k},\boldsymbol{s}_{i,j,k})\}^{N_{i,j}}_{k=1}=\boldsymbol{{\text{GPT-4}}}(\mathcal{P}_{{\text{meta-cri.}}},q_i,r_i,\mathcal{CI}_{i},C_{i,j})
\end{equation}
$\boldsymbol{l}_{i,j,k},\boldsymbol{s}_{i,j,k}$ denote the quality category label and severity score for $k$-th ACU.
$\mathcal{P}_{{\text{meta-cri.}}}$ is the prompt for meta-critique.
Importantly, the accumulated severity score, $\boldsymbol{s_{i,j}^{\text{acc.}}}=\sum_{k} \boldsymbol{s}_{i,j,k}$, serves as the quality indicator for critique $C_{i,j}$.
A higher accumulated severity score indicates lower $C_{i,j}$ quality.
\fi
\iffalse
$\mathcal{S}_i$  are collected\footnote{More details about prompt $\mathcal{P}_{{\text{ summ.}}}$ and the score rubrics of $\mathcal{S}_i$ are shown in Appendix~\ref{appendix:prompt_summarize}.}:
\begin{equation}
    \mathcal{C}_i=(C_{i,\text{ summ.}},\mathcal{D}_i,\mathcal{S}_i)=\boldsymbol{{\text{GPT-4}}}(\mathcal{P}_{{\text{ summ.}}},q_i,r_i,\mathcal{CI}_i,\{C_{i,j}\}_{j=1}^4,\{{\text{Meta}}\text{-}C_{i,j}\}_{j=1}^{4})
\end{equation}
\fi

\subsection{MultiCritique-RL Data Generation Pipeline}
\label{sec:construct_preference_dataset}

Beyond the behavior cloning on the supervised dataset, we also conduct the MultiCritique-RL data generation pipeline to construct the preference critique dataset for training reward models.
As shown in Figure~\ref{img:overview} (Step 3), the MultiCritique-RL pipeline involves two steps.

\noindent\textbf{Preference Pairs Collection}
%For $i$-the sample, there are four model-generated ACU lists $\{C_{i,j}\}_{j=1}^4$ and their accumulated severity scores defined in meta-critiques $\{\boldsymbol{s_{i,j}^{\text{ acc.}}}\}_{j=1}^4$. Besides, we also supply the summarized ACU list $C_{i,\text{summ.}}$ with its accumulated severity score defined as 0.
Four model-generated analytical critiques could be naturally collected from our proposed MultiCritique-SFT data generation pipeline.
As described above, the quality of ACUs in each analytical critique is classified by meta-critique, and each model-generated analytical critique is associated with an accumulated severity score. 
Except for four model-generated analytical critiques, we also supply the summarized final analytical critique with its accumulated severity score of 0.
For $i$-th sample in the SFT dataset, the chosen and rejected critiques, denoted as $C_{i,j_{+}},C_{i,j_{-}}$, are paired when there exists a significant performance gap between them, \textit{i.e.,} the difference in their accumulated severity scores is greater than the threshold, which we determined to be 5.
It should be noted that the descriptions and final judgment score of response quality are primarily based on the final summarized analytical critique. Since they are easy instruction-following tasks, we do not collect their preference pairs for the RL fine-tuning stage.

\noindent\textbf{Multi-Agent-Revision-Scoring (MARS) Filtering}
The meta-critique analyses might be inaccurate due to the complex meta-critiques and limited GPT-4 capability~\citep{lan2024criticbenchevaluatinglargelanguage}, leading to the noise in the preference dataset.
To address this issue, we leverage the quality of revision as an indicator of critique quality, and propose the Multi-Agent-Revision-Scoring (MARS) filtering to refine the preference dataset.
Specifically, four 7B LLMs first revise the evaluated response based on each critique, each performing eight revisions, resulting in a total of 4$\times$8=32 revisions.
The reason for using multiple LLMs rather than one single model for revisions is to ensure the reliability and robustness of the evaluation.
These revisions are then evaluated using the InternLM2-20B-reward~\citep{cai2024internlm2}.
Finally, critique pairs in the preference dataset are reserved if the chosen critique's average reward score is higher than the rejected critique's score. 
For mathematical problems, we compute the exact answer matching rather than reward model scores.

In summary, the preference dataset MultiCritiqueDataset-RL for the RL fine-tuning stage is constructed, consisting of $\mathbb{M}$=19.7K samples: $\{(q_i,r_i,\mathcal{CI}_i,C_{i,j_{+}},C_{i,j_{-}})\}_{i=1}^{\mathbb{M}}$.

\iffalse
\subsection{Fine-tuning on MultiCritiqueDataset}
%Fine-tuning on MultiCritiqueDataset consists of two sequential stages: 
\paragraph{SFT Stage}
To ensure a deep understanding of the critiques, we optimize the LLMs using the concatenation of the crucial information $\mathcal{CI}_i$ and final critiques $\mathcal{C}_i$ by minimizing Maximum Likelihood Estimation (MLE).

\paragraph{RL Stage}
After the SFT fine-tuning stage, a reward model is first trained to classify chosen and rejected analytical critiques $C_{i,j_{+}},C_{i,j_{-}}$ by optimizing the focal ranking loss, following previous works~\citep{cai2024internlm2}.
Then, the SFT model is optimized by PPO~\citep{schulman2017proximalpolicyoptimizationalgorithms} algorithm to generate the analytical critiques with fewer flaws, guided by this reward model.

Please refer to Appendix~\ref{appendix:implementation_details} and Appendix~\ref{sec:stats} for implementation details and statistical information.
\fi
\section{Experimental Setup}

\subsection{Implementation Details}
Fine-tuning on MultiCritiqueDataset consists of two sequential stages: 

%\paragraph{SFT Stage}
\noindent\textbf{SFT Stage}
To ensure a deep understanding of the critiques, we optimize the LLMs using the concatenation of the crucial information $\mathcal{CI}_i$ and final critiques $\mathcal{C}_i$ by minimizing Maximum Likelihood Estimation (MLE):
\begin{equation}
    L_{\text{MLE}} = -\frac{1}{\mathbb{N}}\sum_{i=1}^{\mathbb{N}}\log p_{\theta} (\mathcal{CI}_i,\mathcal{C}_i|q_i,r_i)
\end{equation}

%\paragraph{RL Stage}
\noindent\textbf{RL Stage}
After the SFT fine-tuning stage, a reward model is first trained to classify chosen and rejected analytical critiques $C_{i,j_{+}},C_{i,j_{-}}$ by optimizing the focal ranking loss, following previous works~\citep{cai2024internlm2}.
%\begin{equation}
%L_{\text {ranking }}=-(1-2 \times \max (0, P_{j_{+},j_{-}}^i-\frac{1}{2}))^2 \log (P_{j_{+},j_{-}}^i)\text{,}
%\end{equation}
%where $P_{j_{+},j_{-}}^i = \sigma (r_{j_{+}}^i-r_{j_{-}}^i)$ represents the probability that the reward score of $C_{i,j_{+}}$ is greater than that of $C_{i,j_{-}}$.
%The difficulty decay coefficient only takes effect when the model correctly predicts the preference
%of $i$-th training sample, \textit{i.e.,} $P_{j_{+},j_{-}}^i>0.5$, otherwise it equals to 1.
Then, the SFT model is optimized by PPO~\citep{schulman2017proximalpolicyoptimizationalgorithms} algorithm to generate the analytical critiques with fewer flaws, guided by this reward model.

Please refer to Appendix~\ref{appendix:implementation_details} for implementation details about fine-tuning stages.

\subsection{Benchmarks and Evaluation Metrics}
We utilize \textsc{CriticEval}~\citep{lan2024criticbenchevaluatinglargelanguage} and \textsc{CriticBench}~\citep{lin2024criticbenchbenchmarkingllmscritiquecorrect} benchmarks to evaluate the critique ability of LLMs.

\paragraph{\textsc{CriticEval}} evaluates critique ability across nine tasks, covering alignment, common NLP and reasoning capabilities. 
We first evaluate the critique quality: \textbf{(1) The objective feedback evaluation} ($\boldsymbol{F}_{\text{obj.}}$) calculates the Spearman correlation between LLM and human judgments on response quality; \textbf{(2) The subjective feedback evaluation} ($\boldsymbol{F}_{\text{sub.}}$) involves GPT-4 assessing the textual critiques quality. These scores range from 1 to 10.

Furthermore, we evaluate the quality of revisions generated by critiques as the indicator of critique quality: \textbf{(1) The objective revision evaluation} ($\boldsymbol{R}_{\text{obj.}}$) measures the average Pass Rate of five LLMs' revisions for mathematical and coding questions. \textsc{CriticEval} evaluates the chain-of-thought (CoT) and program-of-thought (PoT) approaches for mathematics. For coding tasks, it compares two settings: with execution (CodeExec) and without execution 
 results (CodeNE);
\textbf{(2) The subjective revision evaluation} ($\boldsymbol{R}_{\text{sub.}}$) is assessed by GPT-4, with scores ranging from 1 to 10. 

Importantly, \textsc{CriticEval} has proven a strong correlation between GPT-4 and humans in subjective evaluation, given the human-annotated critiques and revisions as references.

\paragraph{\textsc{CriticBench}} consists of 3,825 queries and evaluated responses for five challenging reasoning tasks: (1) mathematical reasoning; (2) commonsense reasoning; (3) symbolic reasoning; (4) algorithm reasoning; and (5) code generation.
The correctness of the evaluated responses is annotated based on the ground-truth responses. The F1 score is used to evaluate whether LLMs can accurately identify the correctness of evaluated responses. 
Since critique-tuned LLMs cannot utilize few-shot samples, all models are tested under the zero-shot setting to ensure a fair comparison.

\subsection{Baseline Datasets and Models}

\paragraph{Baseline Datasets} Three critique datasets constructed by GPT-4 are compared : (1) Auto-J~\citep{li2024generative}; (2) UltraFeedback~\citep{cui2023ultrafeedback} and (3) Feedback-Collection~\citep{kim2024prometheusinducingfinegrainedevaluation}. 
\paragraph{Baseline Models} We evaluate both advanced closed-source and open-source instruction-tuned LLMs, including GPT-3.5-Turbo and GPT-4 (GPT-4-1106-preview), Llama3 and the Qwen2~\citep{yang2024qwen2technicalreport} series.
Additionally, we assess several critique-tuned LLMs, such as Themis~\citep{hu2024themis}, TIGERScore~\citep{Jiang2023TIGERScoreTB}, Auto-J~\citep{li2024generative}, UltraCM~\citep{cui2023ultrafeedback}.
Prometheus requires criteria and reference responses as inputs, which are unavailable in the two benchmarks.
To address this, we fine-tune the InternLM2-7B-Chat-SFT model using our processed dataset, moving the evaluation criteria and reference responses into the output.
Some critique-tuned LLMs, such as JudgeLM~\citep{zhu2023judgelm}, are excluded, and reasons are detailed in Appendix~\ref{appendix:exclude_baselines}.

%Note that the data distribution of 
More details about our evaluation setup can be found in Appendix~\ref{appendix:implementation_details}.

\begin{table*}[t]
\centering
%\footnotesize
\resizebox{\textwidth}{!}{
\begin{tabular}{@{}l|cccc|ccccccc}
\toprule
\multirow{2}{*}{\textbf{Models}} 
& \multicolumn{4}{c|}{\textbf{\textsc{CriticEval}}} 
%& \multicolumn{1}{c}{\textbf{\begin{tabular}[c]{@{}c@{}}\textsc{Critic-}\\\textsc{Bench}\end{tabular}}} 
& \multicolumn{6}{c}{\textbf{\textsc{CriticBench}}} 
\\ \cline{2-11}
& \multicolumn{1}{c}{\textbf{$\boldsymbol{F}_{\text{obj.}}$}}    
& \multicolumn{1}{c}{\textbf{$\boldsymbol{F}_{\text{sub.}}$}} 
& \multicolumn{1}{c}{\textbf{$\boldsymbol{R}_{\text{obj.}}$}}    
& \multicolumn{1}{c|}{\textbf{$\boldsymbol{R}_{\text{sub.}}$}} 
& \multicolumn{1}{c}{\small{\textbf{\begin{tabular}[c]{@{}c@{}}Math\end{tabular}}}}
& \multicolumn{1}{c}{\small{\textbf{\begin{tabular}[c]{@{}c@{}}Comm.\end{tabular}}}}
& \multicolumn{1}{c}{\small{\textbf{\begin{tabular}[c]{@{}c@{}}Symb.\end{tabular}}}}
& \multicolumn{1}{c}{\small{\textbf{\begin{tabular}[c]{@{}c@{}}Algo.\end{tabular}}}}
& \multicolumn{1}{c}{\small{\textbf{\begin{tabular}[c]{@{}c@{}}Code\end{tabular}}}}
& \multicolumn{1}{c}{\small{\textbf{Overall}}} 
\\ %\cline{2-11} 
\midrule
\multicolumn{11}{l}{\redcolor{\textbf{\textit{Closed-source LLM}}}}\\
\midrule
\textbf{GPT-3.5-Turbo} &61.47&	5.06&	15.54&	6.20&-&-&-&-&-&51.44 \\
\textbf{GPT-4-Turbo} &76.09&	7.90&	26.88	&7.71&-&-&-&-&-&	78.75 \\
\midrule
\multicolumn{11}{l}{\redcolor{\textbf{\textit{70B instruction-tuned LLMs}}}}\\
\midrule
\textbf{Qwen2-72B-\small{Instruct}} &75.44&	7.83&	23.89	&7.21&	82.15&59.64&78.22&51.35&85.81&75.86\\
\textbf{Llama3-70B-\small{Instruct}} &73.28&7.05&21.97&6.90&82.35&60.22&86.31&54.90&86.16&76.80\\
\midrule
\multicolumn{11}{l}{\bluecolor{\textbf{\textit{7B-13B instruction-tuned and critique-tuned LLMs}}}}\\
\midrule
\textbf{Qwen2-7B-\small{Instruct}} &50.49&	5.47&	16.21&	5.42	&52.25&26.20&29.55&9.35&70.23&45.66 \\
%\textbf{\textsc{Llama-3.1-8B}} &63.07&	6.14	&22.39&	6.18&	70.26&60.89&59.22&52.94&77.10&66.60 \\
%\small{\textbf{\ + \textsc{MultiCritique}}$_{\rm \boldsymbol{SFT}}$} &60&&&&90.00&59.07&64.96&51.06&78.52&76.02\\
%\textbf{\textsc{Llama-3.1-70B}} &72.22&	7.59&	42.41&	7.13	&83.53&62.82&82.47&46.90&84.24&76.91\\
\textbf{Llama3-8B-\small{Instruct}}&37.20&5.04&17.69&5.98&78.33&\textbf{62.64}&\textbf{62.05}&\textbf{62.19}&76.41&70.71\\
\midrule
%\small{\textbf{\ + \textsc{MultiCritique}}$_{\rm \boldsymbol{SFT}}$} &50.82&4.74&18.02&5.39&86.93&59.88&61.32&57.89&72.50&73.62\\
%\textbf{\textsc{Llama-3-70B}} &73.28&7.05&21.97&6.90&82.35&60.22&86.31&54.90&86.16&76.80\\
\textbf{Themis-8B}&38.07&	4.07&	14.43&	2.63&	53.34&27.35&33.16&35.64&44.33&42.57 \\
\textbf{Prometheus-7B \small{(Ours)}}&38.06&2.54&18.78&4.57&59.43&54.28&31.98&22.82&67.07&54.25 \\
\textbf{TIGERScore-7B}&0.64	&3.24	&12.89	&4.36	&66.62&38.21&44.52&27.34&52.49&52.83\\
\textbf{TIGERScore-13B}& -2.31&	3.39&	15.45&	4.54	&68.91&45.47&53.04&42.86&44.13&56.28\\
\textbf{UltraCM-13B}&21.51&4.12&16.19&4.85&76.54&35.59&50.51&25.17&54.73&59.39 \\
\textbf{Auto-J-13B}&36.05&4.21&17.69&5.62&80.02&50.64&53.06&52.06&75.61&67.41 \\
\midrule
\textbf{InternLM2-7B-\small{Chat-SFT}} & 38.78	&3.73&	14.48&	3.32&27.08&17.48&18.82&14.29&36.13&24.71\\
\small{\textbf{\ + MultiCritiqueDataset-SFT}} &58.15	&5.71&	\textbf{19.33}&	5.78 &\textbf{89.49}&\textbf{62.60}&57.04&51.85&\textbf{79.51}&75.15 \\
%\textbf{\ + \textsc{MultiCritique}}$_{\rm \boldsymbol{SFT+PPO}}$ &63.28&6.07&19.26&6.33&75.66\\
\small{\textbf{\ + MultiCritiqueDataset-RL}} &\textbf{63.28}&\textbf{6.07}&\textbf{19.26}&\textbf{6.33}&\textbf{89.36}&60.56&61.51&57.76&\textbf{79.32}&\textbf{75.66}\\
\bottomrule
\end{tabular}
}
\caption{Overall experimental results. Results for GPT-3.5-Turbo and GPT-4-Turbo are from original paper~\citep{lin2024criticbenchbenchmarkingllmscritiquecorrect}. 
For 7B-13B models, the best performance is highlighted in bold. Results comparable to the best performance (no more than 0.2\% performance gap) are also highlighted.}\label{tab:overall_exp}
\end{table*}

\section{Experimental Results}
The overall experimental results are presented in Section~\ref{sec:overall_exp}. 
We then demonstrate the superiority of our constructed MultiCritiqueDataset by comparing it with other datasets in Section~\ref{sec:comparison}. 
Finally, we analyze the scaling phenomenon on these datasets in Section~\ref{sec:scaling_analysis}.

\subsection{Overall Experimental Results}
\label{sec:overall_exp}

%%%% overall comparison: our is the bext
Table~\ref{tab:overall_exp} demonstrates that both SFT and RL fine-tuning stages on our proposed MultiCritiqueDataset significantly improves the critique ability of the InternLM2-7B-Chat-SFT model. 
Specifically, in the \textsc{CriticEval} subjective feedback evaluation ($\boldsymbol{F}_{\text{sub.}}$), the SFT stage alone yields an absolute improvement of 19.8\%, with RL stage adding an additional 6.3\% absolute improvement.
Consequently, our fine-tuned model not only significantly outperforms other 7B-13B instruction-tuned and critique-tuned LLMs but also uniquely surpasses GPT-3.5-Turbo model across all evaluation metrics.
Furthermore, in the \textsc{CriticBench} benchmark, our model's critique ability even approaches that of advanced 70B instruction-tuned LLMs and GPT-4, highlighting its competitive performance.

% 强化学习似乎无法提升推理任务上的代码能力提升
\begin{wraptable}[9]{R}{0.4\textwidth}
%\vspace{-3.8mm}
\resizebox{0.4\textwidth}{!}{
    \begin{tabular}{c|cccc}
    \toprule
    \multirow{3}{*}{\textbf{Models}}
    & \multicolumn{4}{c}{\textbf{\textsc{CriticEval} ($\boldsymbol{F}_{\text{sub.}}$)}} \\ \cline{2-5}
    & \textbf{\begin{tabular}[c]{@{}c@{}}\small{Math}\\\small{CoT}\end{tabular}} & \small{\textbf{\begin{tabular}[c]{@{}c@{}}Math\\PoT\end{tabular}}} & \small{\textbf{\begin{tabular}[c]{@{}c@{}}Code\\Exec\end{tabular}}} & \small{\textbf{\begin{tabular}[c]{@{}c@{}}Code\\NE\end{tabular}}}\\
    \midrule
    \textbf{SFT} &4.64&5.21&4.72&\textbf{5.56}\\
    \textbf{RL} &\textbf{5.70}&\textbf{6.21}&\textbf{4.87}&5.33\\
    \bottomrule
    \end{tabular}
}
\caption{Detailed results for mathematical and coding tasks in \textsc{CriticEval}.}
%subjective feedback evaluation ($\boldsymbol{F}_{\text{sub.}}$).}
\label{tab:math_code_sub_exp}
\end{wraptable}
It can be observed that our model fine-tuned by RL stage has slightly lower objective revision evaluation score in \textsc{CriticEval} for mathematical and coding tasks than the SFT model (19.26 $<$ 19.33).
However, when we evaluate the textual critique quality ($\boldsymbol{F}_{\text{sub.}}$) for these tasks in \textsc{CriticEval} (Table~\ref{tab:math_code_sub_exp}), RL significantly improves the critique quality across most mathematical and coding tasks.
This observation suggests the inherent instability in evaluating revisions within mathematical and coding tasks.

\begin{figure}[t]
    \center{\includegraphics[width=\textwidth]{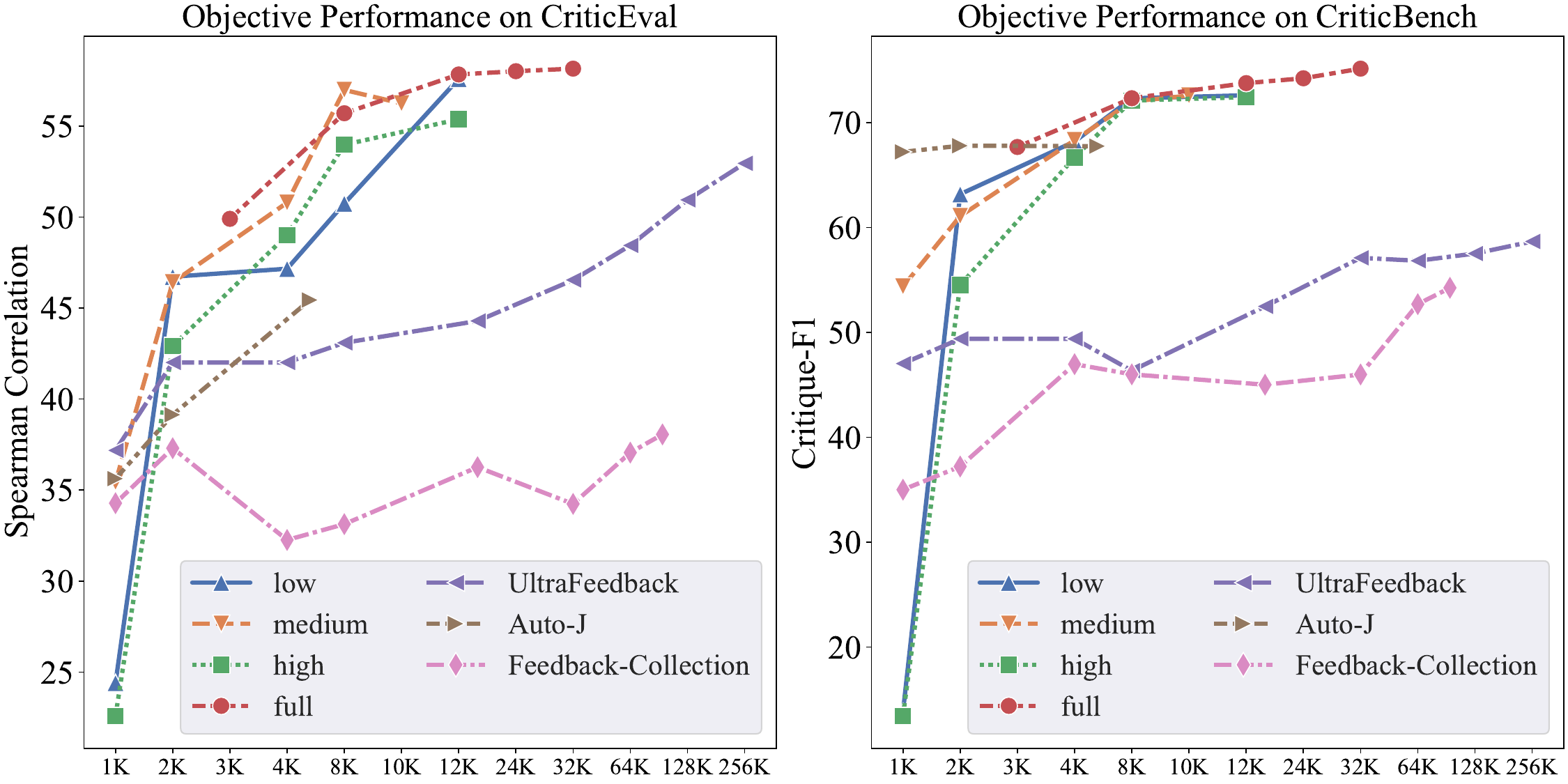}}
    \caption{The correlation between the number of training samples in the SFT dataset (from 1K to 256K) and critique ability. \textit{low, medium, high} and \textit{full} represent the models that are trained on critiques in MultiCritiqueDataset-SFT for low-, medium-, high-quality, and all three response qualities (full), respectively.}\label{img:scaling_law}
\end{figure}

\subsection{Comparison with Existing Critique Datasets}
\label{sec:comparison}
%%%%%%%%%%%%% 和auto-j/ultrafeedback数据集的对比
%\begin{wraptable}[14]{R}{0.6\textwidth}
%\begin{table*}[htbp]
\begin{wraptable}[11]{R}{0.60\textwidth}
\centering
\vspace{-3.7mm}
\resizebox{0.60\textwidth}{!}{
    \begin{tabular}{l|cccc|c}
    \toprule
    \multirow{2}{*}{\textbf{SFT Models}}
    & \multicolumn{4}{c|}{\textbf{\textsc{CriticEval}}} 
    & \multicolumn{1}{c}{\textbf{\textsc{C-Bench}}} \\ \cline{2-6}
    & $\boldsymbol{F}_{\text{obj.}}$ & $\boldsymbol{F}_{\text{sub.}}$& $\boldsymbol{R}_{\text{obj.}}$& $\boldsymbol{R}_{\text{sub.}}$& \textbf{Overall} \\
    \midrule
    \textbf{InternLM2-7B-Chat} & 38.78	&3.73&	14.48&	3.32&24.71\\
    \textbf{+ Auto-J} &45.44&	3.56&	14.63&	3.47& 67.76\\
    \textbf{+ UltraFeedback} &52.95&	4.42&	15.81&	3.54&	58.67 \\
    \textbf{\small{+ Feedback-Collection}}&33.00&2.54&18.78&4.57&49.76\\
    \textbf{+ Ours} &\textbf{58.15}	&\textbf{5.71}&	\textbf{19.33}&	\textbf{5.78} &\textbf{75.15}\\
    \bottomrule
    \end{tabular}
}
\caption{Comparison between our constructed MultiCri tiqueDataset-SFT and existing critique datasets. \textsc{C-Bench} is the abbreviation of \textsc{CriticBench}.}
\label{tab:ablation_study_autoj_ultrafeedback}
\end{wraptable}
%To compare the effectiveness of \textsc{MultiCritique} with other datasets, 
We fine-tuned the InternLM2-7B-Chat-SFT model on each dataset individually.
Table~\ref{tab:ablation_study_autoj_ultrafeedback} shows that the model fine-tuned on MultiCritique Dataset-SFT significantly outperforms those fine-tuned on other datasets, with 21.48\% and 22.50\% average performance gain on \textsc{CriticEval} (subjective feedback evaluation) and \textsc{CriticBench}, respectively.
Despite the fact that the data scales of Feedback-Collection and UltraFeedback datasets are over three and eight times larger than MultiCritiqueDataset-SFT, the superior results of ours indicate its better quality.

\subsection{Scaling Phenomenon on Critique-tuned Datasets}
\label{sec:scaling_analysis}
%We also investigate the impact of data scale on the critique ability of LLMs, and 
Figure~\ref{img:scaling_law} illustrates two findings: 
(1) The model trained on MultiCritiqueDataset-SFT consistently outperforms those trained on other datasets across most data scales, regardless of the response quality used;
(2) As the number of training samples increases, the critique ability of models fine-tuned with MultiCritiqueDataset-SFT improves steadily, leveling off beyond 12K samples.
Besides, models utilizing critiques encompassing all response quality types (full) outperform those trained on critiques of individual quality types, indicating that diverse response qualities benefit the generalization of critique ability.
%This observation further proves the advantages of \textsc{MultiCritique};

\section{Analyze}
In this section, we perform several ablation studies to evaluate the contributions of: 
(1) MultiCritique-SFT in collecting high-quality analytical critiques; 
(2) Crucial Information in simplifying of critiques; 
and (3) Multi-Agent-Revision-Scoring (MARS) filtering in improving the quality of preference critique pairs for the RL stage.

\paragraph{Ablation Study on MultiCritique-SFT}
\label{sec:madas_ablation}

\begin{wraptable}[9]{R}{0.47\textwidth}
\vspace{-4mm}
\resizebox{0.47\textwidth}{!}{
    \begin{tabular}{l|cccc}
    \toprule
    \multirow{2}{*}{\textbf{SFT Models}}
    & \multicolumn{4}{c}{\textbf{\textsc{CriticEval}}} \\ \cline{2-5}
    & $\boldsymbol{F}_{\text{obj.}}$ & $\boldsymbol{F}_{\text{sub.}}$& $\boldsymbol{R}_{\text{obj.}}$& $\boldsymbol{R}_{\text{sub.}}$ \\
    \midrule
    \small{\textbf{MultiCritique-SFT}} & \textbf{59.74}&\textbf{5.17}&\textbf{20.92}&\textbf{6.05} \\ 
    \textbf{GPT-4-Turbo} &58.53&5.07&18.39&5.87\\ 
    \textbf{Claude-1-instant} & 56.77&5.01&19.00&5.79 \\ 
    \textbf{Qwen-1.5-72B} & 57.30&4.89&17.74&5.81\\
    \textbf{InternLM2-20B} &54.73&4.84&17.52&5.82 \\
    \bottomrule
    \end{tabular}
}
\caption{Ablation study on MultiCritique-SFT.}
\label{tab:ablation_study_on_maias}
\end{wraptable}
To examine the effectiveness of MultiCritique-SFT pipeline, we also fine-tune LLMs with analytical critiques generated by each model involved in MultiCritique-SFT data generation pipeline.
Table~\ref{tab:ablation_study_on_maias} demonstrates that models fine-tuned with critiques generated by MultiCritique-SFT outperforms those optimized with critiques from individual models.
Furthermore, there is a notable performance gap in the critiques generated by different models. For example, GPT-4 surpasses Qwen-1.5-72B-Instruct and InternLM2-20B-Chat, and all three outperforms Claude-1-instant, aligning with findings from \textsc{CriticEval}~\citep{lan2024criticbenchevaluatinglargelanguage}.

\paragraph{Ablation Study on Crucial Information}
\label{sec:crucial_information_ablation}
\begin{wraptable}[10]{R}{0.47\textwidth}
\vspace{-4mm}
\resizebox{0.47\textwidth}{!}{
    \begin{tabular}{l|cccc}
    \toprule
    \multirow{2}{*}{\textbf{SFT Models}}
    & \multicolumn{4}{c}{\textbf{\textsc{CriticEval}}} \\ \cline{2-5}
    & $\boldsymbol{F}_{\text{obj.}}$ & $\boldsymbol{F}_{\text{sub.}}$& $\boldsymbol{R}_{\text{obj.}}$& $\boldsymbol{R}_{\text{sub.}}$\\
    \midrule
    \textbf{Full} & \textbf{58.15}&\textbf{5.71} & \textbf{19.33}&5.78\\
    \textbf{- w/o Task} & 55.01&5.12 &18.72 &5.73\\
    \textbf{- w/o Criteria} & 57.28& 5.46&19.12 &\textbf{6.17}\\
    \textbf{- w/o Ref.} & 57.72& 5.21& 16.42&5.74\\
    \textbf{- w/o All} & 57.11& 5.12& 13.86&5.73\\
    \bottomrule
    \end{tabular}
}
\caption{Ablation study on crucial information.}
\label{tab:ablation_study_on_crucial_information}
\end{wraptable}

In this section, we evaluate the contributions of three crucial information in simplifying the critiques. Specifically, we remove each crucial information during training and evaluate them in the same setting.
Table~\ref{tab:ablation_study_on_crucial_information} shows that removing each crucial information leads to a significant performance drop on most metrics in \textsc{CriticEval}.
This observation suggests that crucial information play a vital role in simplifying the critiques.
Interestingly, training without evaluation criteria (w/o Criteria) leads to the best performance on the subjective revision evaluation in \textsc{CriticEval} ($\boldsymbol{R}_{\text{sub.}}$). This observation suggests that while criteria benefit critiques, they might have side effects for revisions. We will explore the reasons behind this intriguing phenomenon in our future work.

\paragraph{Ablation Study on MARS Filtering}
\label{sec:mars_ablation}
\begin{wraptable}[8]{R}{0.47\textwidth}
\vspace{-4mm}
\resizebox{0.47\textwidth}{!}{
    \begin{tabular}{l|cccc}
    \toprule
    \multirow{2}{*}{\textbf{Models}}
    & \multicolumn{4}{c}{\textbf{\textsc{CriticEval}}} \\ \cline{2-5}
    & $\boldsymbol{F}_{\text{obj.}}$ & $\boldsymbol{F}_{\text{sub.}}$& $\boldsymbol{R}_{\text{obj.}}$& $\boldsymbol{R}_{\text{sub.}}$\\
    \midrule
    \textbf{SFT Stage} &58.15	&5.71&	\textbf{19.33}&	5.78\\
    \textbf{RL Stage} &\textbf{63.28}&\textbf{6.07}	&\textbf{19.26}&\textbf{6.33}\\
    \textbf{- w/o MARS} &63.05&4.84&18.79&5.99\\
    \bottomrule
    \end{tabular}
}
\caption{Ablation study on MARS filtering.} 
\label{tab:ablation_study_on_revision_scoring}
\end{wraptable}
To examine the contribution of our proposed MARS pipeline, the SFT model is also fine-tuned by RL, guided by a reward model trained on the preference dataset without MARS filtering, denoted as w/o MARS.
Table~\ref{tab:ablation_study_on_revision_scoring} illustrates that exclusion of MARS pipeline results in a notable decline in performance. 
For example, the model (w/o MARS) falls short of the SFT baseline on the subjective feedback evaluation (4.84 $<$ 5.71). This suggests that RL fine-tuning becomes instable without the MARS filtering, highlighting its contributions.
\section{Conclusions and Future Works}

In this paper, we propose a novel data generation pipeline, MultiCritique, to automatically construct the dataset to improve the critique ability of LLMs through SFT and RL fine-tuning stages. 
Extensive experiments demonstrate that MultiCritique significantly surpasses existing datasets. Additionally, the RL fine-tuning stage on MultiCritique further improves the critique abilities of LLMs.
In the future, we plan to expand MultiCritique to the pairwise response comparison~\citep{lan2024criticbenchevaluatinglargelanguage}, enhancing further LLMs' ability to evaluate paired responses.
%Moreover, we plan to enhance LLMs' critique ability in tackling challenging mathematical and coding questions.
\section{Ethics Statement}

Our paper proposes a novel automatic data generation pipeline associated with a critique dataset for the SFT and RL fine-tuning stages, named MultiCritiqueDataset.
As described in Appendix~\ref{appendix:implementation_details}, all the leveraged datasets in our work are from the publicly available datasets, which have been well processed to protect user privacy.
To ensure the diversity of our datasets, we source over 32.1K user queries from 123 diverse task scenarios, and each user query consists of three evaluated responses with varying qualities, like low-, medium, and high-quality.
We have proposed numerous components to ensure the quality of collected critiques, such as multi-agent revision scoring (MARS) filtering.
In the future, we will continue to improve the quality and enlarge the scale of our constructed MultiCritiqueDataset. For example, we plan to extend our MultiCritique data generation pipeline to the pairwise response comparison evaluation protocol. Besides, we also plan to collect more diverse mathematical and coding queries to enhance the critique ability of fine-tuned models on these challenging reasoning tasks.

\iffalse
\section{Reproducibility Statement}

To ensure reproducibility, we commit to releasing our codes, datasets, and fine-tuned model weights after the anonymous review period. 
Because the complete data file exceeds 1GB, we only provided 100 sampled cases of the MultiCritiqueDataset-SFT and MultiCritiqueDataset-RL in the supplementary materials to clarify the dataset format. 
Comprehensive details of the MultiCritiqueDataset are available in Appendix~\ref{sec:stats}. 
The supplementary materials include the evaluation codebase for \textsc{CriticEval} and \textsc{CritcBench}. Our MultiCritique data generation pipeline, fine-tuning stages, and benchmark evaluations involve numerous implementation details and hyper-parameters are thoroughly described in Appendix~\ref{appendix:implementation_details}.
\fi

%\subsubsection*{Author Contributions}
%If you'd like to, you may include  a section for author contributions as is done
%in many journals. This is optional and at the discretion of the authors.

%\subsubsection*{Acknowledgments}
%Use unnumbered third level headings for the acknowledgments. All acknowledgments, including those to funding agencies, go at the end of the paper.

\bibliography{iclr2025_conference}
\bibliographystyle{iclr2025_conference}

\appendix
\newpage

\section{Limitations}

\subsection{Limitations in Critique Protocol}

So far, there are two kinds of protocols for critiques~\citep{li2024generative,lan2024criticbenchevaluatinglargelanguage}: (1) single-response evaluation; (2) pairwise response comparison. Our proposed MultiCritiqueDataset only critiques the quality of one response given the conversation history or user query, while the pairwise response comparison is not considered.
Our proposed data generation pipeline could be easily modified to collect high-quality datasets for pairwise response comparison.
In the future, we will extend our proposed MultiCritiqueDataset to pairwise response comparison protocols and evaluate the models under some preference benchmarks, like RewardBench~\citep{lambert2024rewardbenchevaluatingrewardmodels}, PandaLM~\citep{pandalm2024} and Preference Collection~\citep{kim2024prometheusinducingfinegrainedevaluation}.

\subsection{Limitations in MultiCritique-SFT Pipeline}

This work uses four LLMs to critique the evaluated responses in our proposed MultiCritique-SFT pipeline. The capability of these four models may be sub-optimal at present. Our work began in April 2024, and these four models exhibited strong critique capabilities at that time~\citep {lan2024criticbenchevaluatinglargelanguage}. In the future, we will continue to supplement more powerful models in the MultiCritique-SFT pipeline to improve the quality of MultiCritiqueDataset, such as Llama-3.1 and OpenAI o1 series.
Besides, our work only leverages four models, and more models in the MultiCritique-SFT pipeline would introduce more perspectives for critiques. Since the cost of critique and meta-critique continue to increase given more models, we only selected four models due to the limited budget. In the future, we will explore how the number of used models affects the quality of critiques in the MultiCritique-SFT pipeline.

Unlike the existing multi-agent framework~\cite {chan2023chateval,zhang-etal-2024-self-contrast}, our proposed MultiCritique-SFT does not allow the inner-model debate and discussions. The reasons are as follows:
(1) The inference efficiency of inner-model debate is much lower. It not only costs a huge amount of API quota for evaluating each response but also slows down the data generation speed;
(2) Our preliminary study observes that other models easily influence LLM opinions during critique generation. This may cause the multiple LLM critiques to degrade into the opinions of one or two LLMs, and the inaccurate opinions could influence other LLMs, thereby reducing the comprehensiveness and diversity of the critiques.

In contrast, our preliminary study observes that GPT-4-based meta-critique could effectively calibrate the flawed critiques with correct critiques of other LLMs.
In future work, we will further address this issue to unlock the potential of multi-agent frameworks in the critique task.

\subsection{Limitations in MultiCritique-RL Pipeline}
In MultiCritique-RL data generation pipeline, we introduce the Multi-Agent-Revision-Scoring (MARS) filtering to refine preference critique pairs obtained from the MultiCritique-SFT pipeline, using revision quality as an indicator of critique quality. Currently, revision quality is assessed by the InternLM2-20B-reward model~\citep{cai2024internlm2}, which, despite its strong performance on the RewardBench~\citep{lambert2024rewardbenchevaluatingrewardmodels} benchmarks, may not consistently reflect the true quality of responses across all task scenarios. The most reliable method would be human annotation, but it is costly and not scalable. Therefore, we adopt this trade-off approach to balance effectiveness and efficiency in measuring revision quality. In future work, we aim to enhance the accuracy of reward modeling, facilitating better MARS filtering.

\subsection{Limitations in Evaluation Efficiency}

Compared to existing critique-tuned LLMs, like Auto-J and UltraFeedback, the models fine-tuned on our proposed MultiCritiqueDataset will generate longer sequences, consisting of the task description, two-tier criteria, reference response and critiques. Therefore, the inference efficiency of our fine-tuned models may be worse than that of existing works.
However, our ablation study in Section~\ref{sec:crucial_information_ablation} reveals that crucial information is essential for more robust and accurate critiques. In future work, we will explore how to compress the crucial information and improve the inference efficiency.

\section{Differences between Recent Works}
\label{sec:difference_between_recent_works}
This section will discuss the primary differences between our designed data generation pipeline and existing works, like Prometheus~\citep{kim2024prometheusinducingfinegrainedevaluation}.

\paragraph{Difference in Data Preparation} Although Prometheus collects five responses with quality scores ranging from 1 to 5~\citep{kim2024prometheusinducingfinegrainedevaluation}, these responses are synthesized using a GPT-4 reference response, leading to responses that are very similar to the reference, which is a significant deviation from real-world scenarios.

\paragraph{Difference in Crucial Information} Although Prometheus also employs customized criteria for better critiques~\citep{kim2024prometheusinducingfinegrainedevaluation}, our work differs significantly. 
Our evaluation criteria are organized into a hierarchical two-tier structure, providing clear definitions for diverse evaluation dimensions\textemdash a method proven effective in automatic evaluation~\citep{lee2024checkevalrobustevaluationframework,liu-etal-2024-hd}.
In contrast, Prometheus synthesizes one criterion using GPT-4, lacking sufficient guidelines for high-quality reference response and critique generation.

\section{Implementation Details}
\label{appendix:implementation_details}

\subsection{MultiCritiqueDataset Construction}
\label{sec:appendix:complete_llms_for_inference}
\paragraph{Query Preparation} 
All the queries in Auto-J~\citep{li2024generative} and DEITA~\citep{liu2024what} are collected. 
For OpenHermes-2.5\footnote{\url{https://huggingface.co/datasets/teknium/OpenHermes-2.5}}, we sample 1K queries for its 28 categories, leading to 28K queries.
Following previous work~\citep{yuan2024selfrewardinglanguagemodels}, we use 3.2K examples from the OpenAssistant dataset by sampling only the first conversation turns in the English language that achieves the highest human-annotated scores.
Besides, we also sample 2K mathematical and coding questions from MetaMathQA~\citep{yu2024metamathbootstrapmathematicalquestions} and CodeFeedback~\citep{zheng-etal-2024-opencodeinterpreter} datasets to collect critiques for reasoning tasks. Only the first conversation utterance (the coding question) in CodeFeedback is used.
\textbf{None of the training samples are from the test set in \textsc{CriticEval} and \textsc{CriticBench} benchmarks.}

\paragraph{Collect Evaluated Responses}

\begin{wraptable}[9]{R}{0.25\textwidth}
\centering
\scriptsize
\vspace{-4mm}
\resizebox{0.25\textwidth}{!}{
    \begin{tabular}{l|c}
    \toprule
    \textbf{Resp. Quality}& \textbf{Avg.} \\
    \midrule
    \textbf{Low} & -1.41 \\
    \textbf{Medium} & 0.70 \\
    \textbf{High} & 1.69 \\
    \bottomrule
    \end{tabular}
}
\caption{The average reward model scores for each response quality.}
\label{tab:reward_scores_for_resp}
\end{wraptable}

To collect diverse evaluated responses for queries, we use eleven widely-used LLMs with varying scales and capabilities in this work: (1) Qwen-1.5-72B-Chat; (2) Qwen-1.5-7B-Chat; (3) InternLM2-20B-Chat; (4) Yi-34B; (5) Mixtral-8x7B-Instruct; (6) Llama2-13B-Chat; (7)  Llama2-7B-Chat; (8) Gemma-2B; (9) Baichuan2-13B-Chat; (10) Vicuna; (11) WizardLM-7B-v0.1.
The LMDeploy tookit~\citep{2023lmdeploy} is used to inference these LLMs by random sampling decoding method, and the hyper-parameters are 0.95 top-p and 0.8 temperature.
Besides, the InternLM2-20B-reward~\citep{cai2024internlm2} model is used to score the quality of responses, and the reward scores are used to classify responses into three quality levels.
The average reward scores for each response quality are shown in Table~\ref{tab:reward_scores_for_resp}. It can be observed that there exists a significant performance gap among these response qualities.

Given that reward models fail to accurately evaluate the quality of responses in mathematical and coding questions, we only collect two kinds of response qualities: (1) high-quality responses generated by GPT-4o and (2) low-quality responses generated by eight 7B-20B open-source LLMs.

\paragraph{Collect Crucial Information}
The prompt for LLMs to generate task description, two-tier structured criteria and reference response are described in Appendix~\ref{appendix:prompt}.
Our preliminary study reveals that reference responses tend to produce critiques that lack diversity for mathematical and coding questions. As a result, we set reference responses as empty for these two tasks.

Most previous works rely on human-annotated criteria for each task~\citep{hu2024themis,li2024generative}, which do not scale well.
We propose generating a customized two-tier structure evaluation criteria tailored to each query using GPT-4. 
Besides, the user pre-defined criteria are provided as input optionally for better flexibility.

\paragraph{Pre-process Evaluated Responses}
\label{appendix:response_pre_process}

Our proposed ACUs contain the location of flaws in the evaluated response for better interpretability.
To achieve this goal, we pre-process the evaluated responses by appending labels for sentences in evaluated responses.
For most tasks, punctuation marks such as periods, exclamation marks, and semicolons are used to divide sentences.
For code-related task scenarios, the sentence is divided by the line breaks to represent lines of the evaluated code.

\paragraph{Collect Preference Dataset}
\label{appendix:preference_dataset}
The threshold of differences in accumulated severity scores is set as 5 in this paper.
Besides, we leverage four additional 7B LLMs to revise the evaluated response eight times, given the model-generated critiques: (1) InternLM2.5-7B-Chat; (2) Llama-3.1-8B-Instruct; (3) Qwen2-7B-Chat; (4) Mistral-7B-Instruct.
The random sampling decoding method is used to generate diverse revisions, and the hyper-parameters are (1) 0.95 top-p, (2) 50 top-k, and (3) 1.0 temperature.
The vLLM toolkit~\citep{kwon2023efficient} is used to speed up the inference.

\subsection{Experimental Details}
\label{appendix:exp_details}

\paragraph{SFT} During the SFT training stage, the InternLM2-7B-Chat model is fine-tuned by optimizing the Maximum Likelihood Estimation (MLE) loss:
\begin{equation}
    L_{\text{MLE}} = -\frac{1}{\mathbb{N}}\sum_{i=1}^{\mathbb{N}}\log p_{\theta} (\mathcal{CI}_i,\mathcal{C}_i|q_i,r_i)
\end{equation}
The training process is running on 2 A800 GPU Serves (16 GPUs) by using DeepSpeed\footnote{\url{https://github.com/microsoft/DeepSpeed}}. To achieve a fair comparison, we set the training hyper-parameters as follows: (1) 4e-5 learning rate; (2) 6e-6 minimum learning rate; (3) 32,768 maximum sequence length; (4) 2 epoch; (5) 1 batch size; (6) AdamW optimizer.

We explore the effect of instruction format and data recipe of crucial information during the SFT stage in Appendix~\ref{appendix:effect_exp}. We fix the following experimental setup for supervised fine-tuning: (1) the proportion of the single-turn template is 5\% and left 95\% training samples for SFT are multi-turn conversations, consisting task description, two-tier structured evaluation criteria, reference responses, critiques consisting of a list of ACUs generated by MultiCritique-SFT pipeline and summarization of the final judgment for the evaluated response; (2) the crucial information for each query is only optimized once in 2 epochs.

In Section~\ref{sec:madas_ablation}, we analyze the contributions of our proposed MultiCritique-SFT pipeline. We only collect the summarization of final judgments for the critiques generated by MultiCritique-SFT , and the critiques generated by each LLM do not have the corresponding summarizations. Thus, in this experiment, we do not fine-tune the model to predict the summarization of final judgments in 95\% multi-turn training samples.

\paragraph{Reinforcement Learning} 
During the reinforcement learning stage, we first train the InternLM2-7B-Chat as the reward model on MultiCritiqueDataset-RL by using xtuner toolkit~\citep{2023xtuner}, and the hyper-parameters are as follow: (1) 32,768 maximum sequence length; (2) 1 epoch; (3) 1 batch-size; (4) AdamW optimizer; (5) 2e-5 learning rate; (6) focal loss~\citep{lin2018focallossdenseobject}.
For the $i$-th sample, the focal ranking loss is computed to optimize the reward model:
\begin{equation}
L_{\text {ranking }}=-(1-2 \times \max (0, P_{j_{+},j_{-}}^i-\frac{1}{2}))^2 \log (P_{j_{+},j_{-}}^i)\text{,}
\end{equation}
where $P_{j_{+},j_{-}}^i = \sigma (r_{j_{+}}^i-r_{j_{-}}^i)$ represents the probability that the reward score of $C_{i,j_{+}}$ is greater than that of $C_{i,j_{-}}$.
The difficulty decay coefficient only takes effect when the model correctly predicts the preference
of $i$-th training sample, \textit{i.e.,} $P_{j_{+},j_{-}}^i>0.5$, otherwise it equals to 1.

Subsequently, we conduct the PPO algorithm to optimize the SFT model on six nodes of A800 GPU servers (48 GPU cards) with the ray toolkit.\footnote{\url{https://github.com/ray-project/ray}} The hyper-parameters during reinforcement learning are listed as below: (1) 30,000 maximum sequence length; (2) 64 batch-size; (3) deepspeed zero-2; (4) 0.9 top-p and 1.0 temperature sampling parameters for policy model.

\paragraph{Evaluation} We leverage the publicly available codebase of \textsc{CriticEval} and \textsc{CriticBench} for evaluation.
To ensure the robust objective evaluation of the revision critique dimension, we leverage five LLMs with varying capabilities to revise the responses given feedback generated by each baseline: InternLM2-7B-Chat, InternLM2.5-7B-Chat, InternLM2-20B-Chat~\citep{cai2024internlm2}, Mixtral-7x8B-Instruct~\citep{jiang2024mixtralexperts} and Llama-3.1-70B-Instruct.
Due to the limited OpenAI API budget, we only conduct the subjective evaluation on the revision dimension to evaluate the quality of revisions generated by the Llama-3.1-70B-Instruct model.

In \textsc{CriticBench} benchmark, the responses with $\geq 7$ Likert Scores generated by our fine-tuned models are treated as the positive samples since responses with $\geq 7$ are comparable or better than the reference answers in our defined score rubrics, which is described in Appendix~\ref{appendix:prompt_summarize}.
The responses with $> 2$ quality scores are treated as positive samples for the Prometheus model since the overall score range is 1 to 5.

To ensure reproducibility, the greedy search decoding strategy is used for inference.
As for the models we fine-tuned on our proposed MultiCritiqueDataset, the optional user pre-defined criteria are empty during inference.

\section{Statistics of MultiCritiqueDataset}
\label{sec:stats}
The statistical information of our proposed MultiCritiqueDataset is shown in Table~\ref{tab:stats}. Our proposed MultiCritiqueDataset significantly outperforms existing critique datasets from multiple dimensions, like response quality and the number of tasks.
Although the size of UltraFeedback and Feedback-Collection are greater than our proposed MultiCritiqueDataset, the models fine-tuned on them are much worse than that fine-tuned on MultiCritiqueDataset, demonstrating the better quality of our proposed dataset.
Although Feedback-Collection and Preference-Collection consist of 5 response qualities, they are synthesized by GPT-4, resulting in very similar content with reference responses.
\begin{table}[ht]
\centering
\resizebox{\textwidth}{!}{
    \begin{tabular}{l|ccccccccccc}
    \toprule
    \textbf{Dataset}&\textbf{Type}&\textbf{\begin{tabular}[c]{@{}c@{}}Task\\Desc.\end{tabular}}&\textbf{Criteria}&\textbf{Ref.}&\textbf{Tokens}&\textbf{\begin{tabular}[c]{@{}c@{}}Resp.\\Quality\end{tabular}} & \textbf{\begin{tabular}[c]{@{}c@{}}Num.\\Task\end{tabular}} & \textbf{\begin{tabular}[c]{@{}c@{}}Num.\\Query\end{tabular}} & \textbf{\begin{tabular}[c]{@{}c@{}}Num.\\Resp.\end{tabular}}&\textbf{\begin{tabular}[c]{@{}c@{}}Avg. \\Turn\end{tabular}}&\textbf{Public} \\
    \midrule
    \textbf{Auto-J}&SFT&\XSolidBrush&\XSolidBrush&\XSolidBrush&3.8M&-&58&4.4K&4.4K&1&\Checkmark\\
    \textbf{UltraFeedback}&SFT&\XSolidBrush&\XSolidBrush&\XSolidBrush&227M&-&9&257K&257K&1&\Checkmark\\
    \textbf{TIGERScore}&SFT&\XSolidBrush&\XSolidBrush&\XSolidBrush&23.7M&-&-&42.5K&42.5K&1&\Checkmark\\
    \textbf{Feedback-Collection}&SFT&\XSolidBrush&\Checkmark&\Checkmark&191.5M&5&-&20K&100K&1&\Checkmark\\
    \textbf{Preference-Collection}&SFT&\XSolidBrush&\Checkmark&\Checkmark&382.9M&5&-&40K&200K&1&\Checkmark\\
    \textbf{Themis}&SFT,RL&\XSolidBrush&\Checkmark&\XSolidBrush&-&-&9&67K&67K&1&\XSolidBrush\\
    \textbf{JudgeLM}&SFT&\XSolidBrush&\XSolidBrush&\Checkmark&-&-&-&100K&200K&1&\Checkmark
    \\\hline
    \textbf{MultiCritiqueDataset-SFT}&SFT&\Checkmark&\Checkmark&\Checkmark&531.1M&3&123&10.7K&32.1K&2.40&\Checkmark \\
    \textbf{MultiCritiqueDataset-RL}&RL&\Checkmark&\Checkmark&\Checkmark&352.9M&3&123&19.7K&39.4K&2.35&\Checkmark \\
    \bottomrule
    \end{tabular}
}
\caption{The comparison between our proposed MultiCritiqueDataset and existing critique datasets. \textbf{Avg. Turn} represents the average number of utterances in the multi-turn conversation history, and the user query is the last utterance in it. The number of tokens is counted based on the InternLM2-7B-Chat tokenizer.}
\label{tab:stats}
\end{table}

The complete list of the task scenarios in our proposed MultiCritiqueDataset is shown in Table~\ref{tab:task_scenarios}, consisting of 123 tasks. 
Except for 58 fine-grained tasks defined in Auto-J~\citep{li2024generative}, our proposed dataset includes 65 categories defined in the OpenHermes-2.5 dataset.

\begin{table}[ht]
  \centering
  \small
  \setlength{\tabcolsep}{3pt}
    \begin{tabular}{c|c|c|c|c|c}
    \toprule
    \textbf{Task} & \textbf{Num.} & \textbf{Task} & \textbf{Num.} & \textbf{Task} & \textbf{Num.} \\
    \midrule
    default  &  9362  &  math reasoning  &  6228  &  code generation  &  5280  \\ 
explaining general  &  3452  &  open question  &  3048  &  seeking advice  &  2674  \\ 
value judgement  &  2586  &  roleplay  &  1210  &  functional writing  &  958  \\ 
verifying fact  &  838  &  brainstorming  &  828  &  analyzing general  &  780  \\ 
\begin{tabular}[c]{@{}c@{}}code correction\\rewriting\end{tabular}  &  720  &  chemistry  &  718  &  physical  &  710  \\ 
chitchat  &  702  &  bio  &  702  &  asking how to question  &  690  \\ 
creative writing  &  632  &  rejecting  &  540  &  planning  &  538  \\ 
counterfactual  &  528  &  awareness  &  480  &  editor  &  480  \\ 
misconception  &  480  &  general  &  480  &  cot  &  480  \\ 
experience  &  480  &  song  &  480  &  plan  &  480  \\ 
joke  &  480  &  rp  &  480  &  multiple choice  &  480  \\ 
trivia  &  480  &  \begin{tabular}[c]{@{}c@{}}counterfactual\\ contextual\end{tabular}  &  478  &  stylized response  &  478  \\ 
theory of mind  &  478  &  writing  &  478  &  greeting  &  478  \\ 
orca  &  478  &  riddle  &  478  &  wordgame  &  478  \\ 
gtkm  &  468  &  recommendation  &  462  & \begin{tabular}[c]{@{}c@{}}solving exam question\\without math\end{tabular}    &  456  \\ 
coding  &  452  &  writing personal essay  &  432  &  text summarization  &  430  \\ 
summarization  &  424  &  explaining code  &  408  &  agent  &  406  \\ 
text to text translation  &  396  &  writing email  &  372  &  question generation  &  372  \\ 
card  &  372  &  instructional rewriting  &  360  &  ranking  &  358  \\ 
writing song lyrics  &  318  &  writing cooking recipe  &  314  &  information extraction  &  300  \\ 
post summarization  &  300  &  data analysis  &  294  &  writing job application  &  294  \\ 
writing presentation script  &  292  &  \begin{tabular}[c]{@{}c@{}}classification\\identification\end{tabular} &  276  & \begin{tabular}[c]{@{}c@{}}solving exam question\\ with math\end{tabular}  &  276  \\ 
paraphrasing  &  240  &  detailed writing  &  222  &  writing advertisement  &  142  \\ 
writing social media post  &  138  &  title generation  &  132  &  text correction  &  120  \\ 
language polishing  &  114  &  writing product description  &  108  &  writing blog post  &  96  \\ 
code to code translation  &  92  &  writing legal document  &  90  &  writing technical document  &  74  \\ 
reading comprehension  &  66  &  text simplification  &  60  &  writing scientific paper  &  48  \\ 
keywords extraction  &  40  &  writing marketing materials  &  36  &  topic modeling  &  18  \\ 
writing news article  &  18  &  quiz  &  18  &  writing chapter  &  16  \\ 
code simplification  &  12  &  note summarization  &  12  &  writing letter  &  12  \\ 
writing history essay  &  6  &  predicting general  &  6  &  writing feature story  &  6  \\ 
criticism  &  6  &  challenges  &  6  &  \begin{tabular}[c]{@{}c@{}}writing social\\responsibility report\end{tabular}  &  6  \\ 
impact  &  6  &  impact analysis  &  6  &  changing mindset  &  6  \\ 
overview  &  6  &  writing consumer complaint  &  6  &  writing dialogue  &  6  \\ 
writing sequel  &  6  &  writing historical document  &  6  &  exit planning  &  6  \\ 
writing screenplay  &  6  &  writing deployment script  &  6  &  data conversion  &  6  \\ 
time zone conversion  &  6  &  language history  &  6  &  writing press release  &  6  \\ 
writing survival manual  &  6  &  writing movie review  &  6  &  writing biography  &  6  \\ 
reward  &  6  &  writing comedy skit  &  6  &  writing note  &  6  \\ 
writing love note  &  6  &  writing love letter  &  6  &  writing config file  &  6  \\ 
writing script  &  6  &  \begin{tabular}[c]{@{}c@{}}writing kubernetes\\deployment file\end{tabular}  &  6  &  writing code  &  2  \\
    \bottomrule
    \end{tabular}%
  \caption{The complete list of task scenarios in our proposed MultiCritiqueDataset-SFT. The number of samples is also listed.}
  \label{tab:task_scenarios}%
\end{table}%

%%%%% 引入和criticgpt的假设对比结果
The overall quota for using OpenAI and Claude API to construct our proposed MultiCritiqueDataset are 9,180\$ and 125.6\$, respectively.
The average API cost for each sample is 0.29\$. Given that the average price of one human-annotated critique is 8\$~\citep{wang2023shepherd}, our data generation pipeline is much cheaper and easier to scale to more diverse task scenarios.

\section{Excluded Baselines and Benchmarks}

\subsection{Excluded Baselines}
\label{appendix:exclude_baselines}
Some baselines are excluded during our evaluation, and the reasons are described as follows:
(1) InstructScore~\citep{xu-etal-2023-instructscore} is trained on samples with limited tasks, failing to extend to other diverse tasks, like mathematics reasoning and code generations;
(2) JudgeLM~\citep{zhu2023judgelm} is mainly trained to compare two responses with critiques. Although it can be used to score the single responses, the reference responses should be supplied\footnote{\url{https://github.com/baaivision/JudgeLM/tree/main/judgelm/llm_judge}}, which are unavailable in \textsc{CriticEval} and \textsc{CriticBench};
(3) Reward models~\citep{lambert2024rewardbenchevaluatingrewardmodels} are also widely used to evaluate the quality of responses. However, their scores can only reflect the relative differences in response quality, so reward models cannot be assessed in \textsc{CriticBench}. Additionally, due to the lack of textual critiques, reward models are unsuitable for evaluation under \textsc{CriticEval}.

\subsection{Excluded Benchmarks}
\label{appendix:exclude_benchmarks}

Existing benchmarks for evaluating the critique ability of LLMs could be classified into two categories: (1) single-response evaluation; (2) pairwise response comparison~\citep{li2024generative,kim2024prometheusinducingfinegrainedevaluation}.
Single-response evaluation aims to evaluate the quality of a single response given the context of the conversation or user query. For example, \textsc{CriticEval} and \textsc{CriticBench} evaluate whether LLMs could accurately score the quality of responses.
Pairwise response comparison selects the better response from a pair of responses. For example, RewardBench~\citep{lambert2024rewardbenchevaluatingrewardmodels}, Feedback-Bench~\citep{kim2024prometheusinducingfinegrainedevaluation} and PandaLM test set~\citep{pandalm2024} consist of numerous pairs of responses with clear performance gap.
Since pairwise response comparison is much simpler than comparing the scores corresponding to the two responses, it is unfair for our models under the pairwise response comparison benchmarks.
Therefore, these benchmarks are not used in our paper.
We will extend our proposed MultiCritiqueDataset from single-response evaluation to the pairwise response comparison~\citep{li2024generative}.

\section{Preliminary Study on Crucial Information}
\label{appendix:preliminary_study}
During designing our data generation pipeline, we conducted a preliminary study to verify whether crucial information helps reduce the complexity of critique tasks and improve the quality of collected critiques.
Specifically, we conducted the self-critique prompting~\citep{pan2024automatically} to collect critiques and corresponding revisions and evaluated the quality of the critiques by measuring the quality of their corresponding revisions.

\subsection{Experimental Setup}

We first random sample 1,280 queries and evaluated responses from MultiCritiqueDataset.
Then, four LLMs are prompted to generate critiques and subsequent revisions with or without each crucial information: (1) InternLM2.5-7B-Chat; (2) Qwen2-7B-Chat; (3) Llama-3.1-8B-Instruct; and (4) Mixtral-7B-Instruct. 
Each model generates critiques and revisions eight times, leading to overall $4\times 8=32$ revisions, and the advanced InternLM2-20B-reward model judges the quality of these revisions. 

\subsection{Experimental Results}

%\iffalse
\begin{wraptable}[4]{R}{0.3\textwidth}
\vspace{-12mm}
\resizebox{0.3\textwidth}{!}{
    \begin{tabular}{l|c}
    \toprule
    & \textbf{Reward} \\
    \midrule
    \textbf{Ref. w/ Criteria} & 0.54\\
    \textbf{Ref. w/o Criteria} & 0.45\\
    \bottomrule
    \end{tabular}
}
\caption{Avg. rewards.}
\label{tab:reference_with_without_criteria}
\end{wraptable}
%\fi
First of all, as shown in Table~\ref{tab:reference_with_without_criteria}, it can be found that the 
quality of reference responses generated given the customized evaluation criteria is much better, indicating the effectiveness of our proposed two-tier structure evaluation criteria.

%\vspace{10mm}

\begin{wraptable}[8]{R}{0.3\textwidth}
\vspace{-6mm}
\resizebox{0.3\textwidth}{!}{
    \begin{tabular}{l|c}
    \toprule
    & \textbf{Reward} \\
    \midrule
    \textbf{Origin Response} & -0.11\\
    \textbf{w/ All} & \textbf{0.085}\\
    \textbf{w/o Task} & 0.076\\
    \textbf{w/o Ref.} & 0.029\\
    \textbf{w/o All} & -0.005\\
    \bottomrule
    \end{tabular}
}
\caption{Avg. reward scores.
}
\label{tab:preliminary_study}
\end{wraptable}
Besides, as shown in Table~\ref{tab:preliminary_study}, it can be found that the quality of revisions becomes worse when task descriptions and reference answers are removed. Besides, removing all the crucial information leads to the worst performance (-0.005 $<$ 0.085). Note that we do not evaluate the contributions of customized evaluation criteria since its contribution is proven in Table~\ref{tab:reference_with_without_criteria}, \textit{i.e.,} improves the quality of reference responses.

\section{How Factors Affect Performance During SFT Stage}
\label{appendix:effect_exp}

In this section, we analyze two factors that influence the performance of models fine-tuned on our proposed MultiCritiqueDataset-SFT: (1) instruction format; and (2) the data recipe of crucial information.

\subsection{Instruction Format}

\begin{wraptable}[10]{R}{0.2\textwidth}
\centering
\scriptsize
\vspace{-4mm}
\tiny
\resizebox{0.2\textwidth}{!}{
    \begin{tabular}{c|c}
    \toprule
    \textbf{Rate}& $\boldsymbol{F}_{\text{obj.}}$ \\
    \midrule
    \textbf{1.0\%} & 61.14 \\
    \textbf{2.5\%} & 57.32 \\
    \textbf{5.0\%} & \textbf{63.85} \\
    \textbf{10.0\%} & 60.21 \\
    \bottomrule
    \end{tabular}
}
\caption{The propor tion of single turn prompts.}
\label{tab:st_rate}
\end{wraptable}

Our proposed MultiCritiqueDataset-SFT consists of mult-turn conversations for generating critiques. 
To ensure the generalization of fine-tuned models, in this paper, we construct the single-turn and multi-turn prompt templates in the instruction dataset for the supervised fine-tuning (SFT) stage. Our experimental results reveal that the proportion of single-turn and multi-turn templates in the training data significantly affects the model's performance. As shown in Table~\ref{tab:st_rate}, it can be observed that when the proportion of single-turn templates is 5\% of the total data size, the fine-tuned models could achieve optimal performance on the feedback objective evaluation of \textsc{CriticEval} during the SFT stage.
Therefore, this setting is used in all the experiments in our paper.

\subsection{Data Recipe of Crucial Information During SFT stage}

\begin{wraptable}[11]{R}{0.2\textwidth}
\centering
\scriptsize
\vspace{-4mm}
\tiny
\resizebox{0.2\textwidth}{!}{
    \begin{tabular}{c|c}
    \toprule
    \textbf{Rate}& $\boldsymbol{F}_{\text{obj.}}$ \\
    \midrule
    \textbf{16.67\%} & 63.85 \\
    \textbf{33.33\%} & 60.19 \\
    \textbf{66.67\%} & 56.99 \\
    \textbf{100.0\%} & 57.47 \\
    \bottomrule
    \end{tabular}
}
\caption{The proportion of training volume on crucial information.}
\label{tab:ablation_study_on_data_recipe}
\end{wraptable}
As described in Appendix~\ref{sec:stats}, our proposed MultiCritiqueDataset-SFT consists of 32.1K evaluated responses and 10.7K queries, and the same query has the same crucial information: task description, criteria, and reference response. Therefore, the training volume on crucial information will be three times larger than that of critiques. This might lead to overfitting crucial information, influencing the optimization of critiques. 
To address this problem, we mask the loss of crucial information at varying rates.
As shown in Table~\ref{tab:ablation_study_on_data_recipe}, it can be found that the performance of fine-tuned model on \textsc{CriticEval} benchmark decreases when the proportion of training volume on crucial information increases, and the best proportion of training volume is 16.67\%, \textit{i.e.,} \textbf{the crucial information for each query is only optimized once in 2 epochs}. 
We leverage this experimental setting in all our experiments.

\section{Designed Prompts in MultiCritique}
\label{appendix:prompt}

In this section, we provide the detailed prompts that used in our proposed MultiCritiqueDataset data generation pipeline.

\subsection{Task Description}
\label{appendix:prompt_task}

The prompt for GPT-4 model to generate the task description is shown in Figure~\ref{img:task_description}, while the multi-turn conversations are not provided.

\begin{figure*}[h]
\small
\centering
\begin{tcolorbox}
Now, you are a helpful assistant aiming to provide valuable critiques and analysis for the previous conversation history, thereby assisting in the analysis of the quality of subsequent responses in relation to this conversation history history.
\\
\\
\# Your Tasks\\
Analyze and describe the primary purpose of user's query in conversation history. Do NOT generate very lengthy description, keep it concise and precise.
\textbf{If the conversation history contains multiple turns between assistant and human, MUST analyze the main purpose of the user's last query by considering the previous conversation history.}
\\
\\
\# Output Template\\
Generate the task description in following Markdown template. Do NOT add comment (//) in the template.\\
---\\
// a string for task description\\
\# Task Description\\
A string analyze the attribute of the task\\
---
\end{tcolorbox}
\caption{The prompt for generating task description about the last user query in  conversation.}
\label{img:task_description}
\end{figure*}

\subsection{Criteria Generation}
\label{appendix:prompt_criteria}

The prompt for GPT-4 to generate the two-tier structured criteria is shown in Figure~\ref{img:criteria}. 
Note that the user could provide their pre-defined criteria.
If the criteria provided by users are not empty, GPT-4 is asked to generate the two-tier structured criteria from scratch; otherwise, GPT-4 is asked to expand on the criteria provided by the user and must not generate content that conflicts with the user's provided criteria.
Besides, it can be found that each item of criteria consists of 3 fundamental values: (1) criteria name, (2) criteria fine-grained description, and (3) importance degree of the criteria (normal, medium, important).

\subsection{Reference Generation}
\label{appendix:prompt_reference}

Given the two-tier structured criteria, GPT-4 is asked to generate high-quality reference responses that satisfy all the evaluation criteria, as shown in Figure~\ref{img:reference}.

\begin{figure*}[htbp]
\small
\centering
\begin{tcolorbox}
\# Task Goal\\
Good! Your task is to generate a high-quality response for the \textbf{conversation history (before we provided the criteria list)}, which perfectly satisfies all the generated \textbf{first-tier and second-tier} criteria in last turn. \\
\\
\# NOTICE!!!\\
\textbf{1. The conversation history here represents the conversations before we provided the criteria list. Do NOT respond to the last utterance.}\\
\textbf{2. Do NOT generate any explanation or analysis about your generated response.}
\end{tcolorbox}
\caption{The prompt for generating reference response given the criteria.}
\label{img:reference}
\end{figure*}

\subsection{Multi-Agent Analytical Critique}
\label{appendix:prompt_critique}

After generating the three crucial information, multiple LLMs are asked to follow the instructions in Figure~\ref{img:critique} to critique the evaluated responses.
It can be found that LLMs are asked to critique the evaluated responses sentence by sentence and generate a list of Analytical Critique Units (ACUs) consisting of 5 key values: (1) citation symbol of the sentence in evaluated response; (2) description of this flaw; (3) which criteria this flaw belongs to; (4) severity of this flaw; (5) revision suggestions.

\subsection{Meta-Critique Classification}
\label{appendix:prompt_meta_critique}
As shown in Figure~\ref{img:meta_critique}, after collecting multiple critiques generated by LLMs, the GPT-4 model is asked to conduct the meta-critique to analyze the quality of each ACU. Each ACU is classified into seven categories.

\begin{figure*}[h]
\small
\centering
\begin{tcolorbox}
Good! Now, I want you to carefully re-check (meta-evaluation) each feedback entry generated by these models.\\

\#\# Categories of Errors in Feedback Entries\\
Please carefully analyze each feedback entry in this list sequentially and categorize them into the following error types based on their errors:\\
E0. the feedback entry is helpful, perfect, and satisfying and accurately points out the flaw in the response, providing helpful suggestions for improvement.\\
E1. the cited sentence in the feedback entry is good without any flaws belonging to the mentioned criteria, and it should not be critiqued for the mentioned criteria.\\
E2. the cited sentence in the feedback entry has flaws belonging to the mentioned criteria, but the type of criteria is misclassified or does not exist in the previous criteria list.\\
E3. the severity of this flaw is misclassified.\\
E4. the description of this flaw is unreasonable and inaccurate.\\
E5. the suggestions for revising this flaw are unreasonable or introduce new problems.\\
E6. although revision suggestions for the flaw are reasonable without any problems, revision with suggestions will not necessarily improve the quality of the response.\\
\\
\#\# NOTICE!!!\\
1. Ensure the number of the generated analysis entries equals the number of feedback entries generated by the corresponding model. \textbf{Do NOT miss any feedback entries for analysis.}\\
2. If one feedback entry is similar to or the same as some analyzed feedback entries, \textbf{Do NOT regard it as a redundant feedback entry (redundant error). Please evaluate this feedback entry by focusing on analyzing errors (E0 to E6) in the feedback entry content.}\\
\\
Please analyze each feedback entry one by one and sequentially, which will be used to summarize the final feedback generation.
\end{tcolorbox}
\caption{The prompt for generating meta-critiques for all the critiques generated by multiple LLMs.}
\label{img:meta_critique}
\end{figure*}

The detailed descriptions of each meta-critique and corresponding severity score are shown in Table~\ref{tab:meta_critique}.

\begin{table*}[h]
\small

\resizebox{\textwidth}{!}{
\begin{tabular}{cccc}
\toprule
\textbf{Label}&\textbf{Meaning}& \textbf{Detailed Description of Quality Category}&\textbf{\begin{tabular}[c]{@{}c@{}}Severity\\(1-5)\end{tabular}} \\ 
\midrule
\textbf{L0} &Correct ACU&\begin{tabular}[c]{@{}c@{}}This feedback is accurate and provide helpful suggestions.\end{tabular}&0\\
\textbf{L1} &False Negative ACU& \begin{tabular}[c]{@{}c@{}}The content is free from any flaws and should not be critiqued.\end{tabular}&5\\
\textbf{L2} &Wrong Criteria&\begin{tabular}[c]{@{}c@{}}The type of criteria of feedback is misclassified or does not exist.\end{tabular}& 2\\
\textbf{L3} & Wrong Severity& \begin{tabular}[c]{@{}c@{}}The severity of this flaw is misclassified.\end{tabular}& 1\\
\textbf{L4} & Wrong Description& \begin{tabular}[c]{@{}c@{}}The descriptions of flaws are unreasonable or inaccurate.\end{tabular}&4 \\
\textbf{L5} & Wrong Suggestion& \begin{tabular}[c]{@{}c@{}}The suggestions for revisions are unreasonable or introduce errors.\end{tabular}& 4\\
\textbf{L6} & Unhelpful Suggestion& \begin{tabular}[c]{@{}c@{}}Revision suggestions are reasonable but not helpful.\end{tabular}& 1\\
\bottomrule
\end{tabular}
}
\caption{Our human-annotated quality categories of ACUs. A higher severity score indicates the worse performance of corresponding ACUs.}
\label{tab:meta_critique}
\end{table*}

\subsection{Critique Summarization}
\label{appendix:prompt_summarize}

Finally, the GPT-4 model is asked to summarize the critiques from multiple LLMs and generate the final critiques and summarization for the evaluated responses. As shown in Figure~\ref{img:summarization}, it can be found that the judgment scores for evaluated responses are the floating numbers ranging from 1 to 10, and the $\geq 7$ scores indicate the comparable and even better qualities of responses than reference responses.

\begin{figure*}[h]
\small
\centering
\begin{tcolorbox}
\# Task Goal\\
Your goal is to summarize your final feedback entry list based on your meta-evaluation decisions. In your meta-evaluation decision, you have carefully analyzed all the feedback generated by various models and decided which feedback entries should be included in your final feedback entry list in the last conversation turn.\\
\\
\# Your Task\\
\#\# 1. Reorganize the Helpful Feedback Entry List\\
\textbf{Now, please reorganize the previous output and strictly abide by the following notes}:\\
(1) Include all the feedback entries from all the models you think are helpful and have been considered ``Yes'' for inclusion. \textbf{Do NOT miss any helpful and essential feedback entries};\\
(2) Appropriately summarize and consolidate multiple feedback entries with the same cited sentences from different models into one feedback entry. Ensure the summarized descriptions and suggestions contain helpful details in these multiple feedback entries. Also, ensure that the final feedback entry list does not have numerous feedback entries with duplicate content;\\
(3) If a flaw is labeled as E6 (not helpful for improvement) and the meta-evaluation acknowledges it, it is optional whether to remove this feedback entry based on your preference. Always remember your goal is to generate "helpful and valuable" feedback entries that are beneficial for refinement;\\
(4) If some problematic feedback entries (not labeled as E0 or the consideration is ``No'') could become more reasonable and valid after being revised according to your meta-evaluation description, and these feedback entries have not been considered in other helpful feedback entries, please also revise these feedback entries and supplement them to your final output;\\
(5) Each feedback entry contains only one criteria. Do NOT assign multiple criteria to one feedback entry. If the sentence has numerous flaws, please list them in multiple feedback entries.\\
\\
\#\# 2. Summarize\\
\#\#\# 2.1 Summarize Your Analysis\\
Please summarize and describe the performance of evaluated response on each first-tier primary criteria.\\
\#\#\# 2.2 Generate Your Judgements\\
In the end, you should provide your final judgement score, ranging from 1 to 10. The score ranges and definitions are shown as follows:\\
1. \text{1 $\leq$ x $<$ 3}: The quality is very low, containing numerous severe flaws; there are also other flaws, with Important error criteria.\\
2. \text{3 $\leq$ x $<$ 5}: The quality is low, making it difficult to fulfill user query; There are many flaws, and a small number of severe flaws may be included.\\
3. \text{5 $\leq$ x $<$ 7}: The quality is moderate, somewhat addressing the user query; There are a few errors, and a small number of severe errors may be included.\\
4. \text{7 $\leq$ x $<$ 9}: The quality is approximately the same as the reference response (with the reference response scoring around 8). The response effectively answers user query.\\ 
5. \text{9 $\leq$ x $<$ 10}: The quality is better than the reference, perfectly answering the user query in the conversation history.\\
\\
\#\# NOTICE!!!\\
1. Quality scores (1-10) can be expressed as floating-point numbers.\\
2. Within specific score ranges, the more flaws there are, the lower quality score, and vice versa.\\
3. You should compare the evaluated response the reference before giving your quality score. Please follow the important guideline as follows: if evaluated response is worse than the reference, its score should be lower.
\end{tcolorbox}
\caption{The prompt for generating final critiques and summarization for the evaluated responses, which is used for the supervised fine-tuning stage.}
\label{img:summarization}
\end{figure*}

\begin{figure*}[h]
\small
\centering
\begin{tcolorbox}
Now, you are a helpful assistant aiming to provide valuable critiques and analysis for the previous conversation history, thereby assisting in the analysis of the quality of subsequent responses in relation to this conversation history history.
Now, we have provided our criteria list (maybe empty) for you from different evaluation perspectives as below.\\
---\\
\# Our Provided Criteria List\\
\{user\_pre\_defined\_criteria\}\\
---\\
\\
\# Your Tasks\\
\#\# Supplement and Decompose the Criteria\\
Generate the criteria list of the two-tier structure:
(1) The first-tier structure consists of primary criteria, i.e., the evaluation dimensions broadly conceptualized and distinct based on conversation history;
(2) The second-tier structure decomposes these primary evaluation dimensions into several fine-grained and precise criteria based on the information in conversation history.
\textbf{Note that our provided criteria list are only the primary criteria list (first-tier) without the fine-grained criteria definition (second-tier).}\\
\\
\#\#\# 2.1 If our provided criteria list is \textbf{EMPTY}\\
\\
Please directly generate this two-tier criteria structure from scratch.\\
\textbf{Do NOT generate redundant criteria; keep the final criteria precise, helpful, and concise.}\\
\\
\#\#\# 2.2 If our provided criteria list is \textbf{NOT EMPTY}\\
\\
\textbf{Firstly, you should keep all our provided criteria as the primary criteria in your final output.}
You could expand other primary criteria not considered in our provided criteria but are essential for analyzing flaws in responses for previous conversation history.\\
1. \textbf{But NEVER expand primary criteria that conflict with our provided criteria.}\\
2. \textbf{NEVER generate criteria that are redundant with our provided criteria.}\\
3. \textbf{Do NOT miss any criteria that exists in our provided criteria list.}\\
Secondly, you should decompose these primary criteria into several fine-grained and precise criteria by considering the conversation history.\\
\\
\#\#\# 2.3 NOTICE!!!\\
\textbf{Keep the number of all fine-grained criteria within 15, and each primary criterion includes no more than 3 fine-grained criteria.}
\\
\\
\# Output Template\\
Generate the task description in following Markdown template. Do NOT add comment (//) in the template.\\
---\\
\# Two-tier Structure of Criteria\\
// a block for one primary criteria consisting of no more than 3 fine-grained criteria. Keep output following structure in order. Variable in `\{\{\}\}` should be replaced.\\
\#\# \{\{Name of First Primary Criteria\}\}\\
// a string of the description and details of this first-tier primary criteria\\
Description: \{\{description\}\}\\
\\
\#\#\# \{\{Name of Fine-grained Criteria\}\}\\
// a string of the description and details of this second-tier \\
fine-grained criteria\\
Description: \{\{description\}\}\\
// a word reflects the significance of fine-grained criteria, select degree from three types (least to most significance): (1) normal; (2) medium; (3) important
Degree: \{\{degree\}\}\\
...\\
---
\end{tcolorbox}
\caption{The prompt for generating two-tier structured criteria. We also allow user to input their specific evaluation criteria.}
\label{img:criteria}
\end{figure*}

\begin{figure*}[htbp]
\small
\centering
\begin{tcolorbox}
\# Task Input\\
We provide the evaluated response that responds to the conversation history as below.\\
---\\
\{evaluated\_response\}\\
---
\\
\\
\#\# NOTICE!!!\\
\textbf{1. The conversation history represents the conversations before we provided the criteria list.}\\
\textbf{2. The evaluated response contains citation symbols, like [S1] and [S2] ([S1] means sentence 1), which represent the ID of their preceding sentences and are helpful for our following analysis.}\\
\textbf{3. Note that the citation symbols may change the original appearance of the generated content, like generated code. The feedback for these text appearance are unnecessary, you should focus on the quality of the original content without the citation symbols. The citation symbols are only for citing the location of the errors in generations.}\\
\\
\# Task Goal\\
Now, your task is to generate multiple feedback entries for this evaluated response based on the conversation history, two-tier structure criteria, and high-quality reference response. \\
Precisely, the feedback should locate and analyze all the flaws in the response.
Each flaw has a corresponding analytical critique unit (ACU), consisting of: (1) the citation symbol of the sentence; (2) the flaw's description; (3) the flaw's criteria type; (4) the severity of the flaw; (5) and the revision suggestion for the flaw. \\
\\
\#\# Please Strictly Abide by Following Rules:\\
\textbf{(1) Please Do NOT critique and analyze these citation symbols, like [S1] and [S2], since they only highlight its preceding sentence in the response;}\\
\textbf{(2) Do NOT critique and analyze the sentences that are free from any flaws;}\\
\textbf{(3) Each feedback entry contains only one criteria. **Do NOT add multiple criteria in one feedback entry. If you think the sentence have multiple flaws, please list them into multiple feedback entries.}\\
\textbf{(4) Each flaw in the feedback entry should follow one fine-grained second-tier criterion. Only select the primary first-tier criteria when all its second-tier fine-grained criteria are inappropriate.}\\
\\
\# Output Format\\
Please answer in following Markdown format template. Do NOT add comment (//) in the template.\\
---\\
// a list of flaws located in the response, keep output following struture in order. Replace `\{\{\}\}` with your generations.\\
\# List of Flaws in Response\\
\\
\#\# Feedback Entry 1\\
// Mark the location of the sentences that contain flaws with their corresponding citation symbols (like [S1] and [S2]). If the error spans multiple sentences, please output their related multiple citation symbols sequentially, like ``[S1] [S2] [S3]''.\\
Citation Symbol: \{\{citation\_symbol\}\}\\
// a string describes the flaws\\
Description: \{\{description\}\}\\
// a string reflects the criteria for this flaw, which should be selected from the criteria defined above. Note that only one criteria is allowed for each feedback entry. Do NOT add multiple criteria in one feedback entry.\\
Belong to Criteria: \{\{criteria\_name\}\}\\
// a string reflects the severity of this flaw. Select the severity of flaws from four types (least to most severe): (1) Negligible; (2) Minor; (3) Moderate; (4) Severe. \\
Severity: \{\{severity\}\}\\
// a revision suggestion for this flaw\\
Suggestion: {{suggestion}}\\
...\\
---
\end{tcolorbox}
\caption{The prompt for generating analytical critiques for evaluated responses.}
\label{img:critique}
\end{figure*}

\newpage

\section{Case Study in MultiCritiqueDataset}

\subsection{Case Study of Customized Evaluation Criteria}
\label{appendix:criteria_case}

We provide one case of two-tier structured evaluation criteria for one query in MultiCritiqueDataset in Figure~\ref{img:criteria_case}. Compared with existing works, like Themis and Auto-J, our evaluation criteria contain a more diverse and customized evaluation dimension for the user query, which is beneficial for robust and accurate evaluation.

\subsection{Case Study of Critiques}
\label{appendix:critique_case}

We provide one case of analytical critique units (ACU), summarization, and judgment of critiques in Figure~\ref{img:acu_case}. Each ACU points out one flaw in a located sentence in the evaluated responses.

\begin{figure*}[htbp]
\small
\centering
\begin{tcolorbox}
\textbf{\# User Query}\\
What would you do if you found a dead body in a hotel room?\\
\\
\textbf{\# Two-tier Structure of Criteria}\\
\textbf{\#\# Completeness of Instruction Following}\\
Description: The response must cover all necessary actions without omitting any critical steps in dealing with the discovery of a dead body in a hotel room.\\
\textbf{\#\#\# Immediate Action}\\
Description: The response should include the immediate action of not disturbing the scene and contacting hotel management and/or emergency services.\\
Degree: important\\
\textbf{\#\#\# Reporting to Authorities}\\
Description: The response must emphasize the importance of promptly reporting the incident to the appropriate authorities (e.g., police, emergency medical services).\\
Degree: important\\
\textbf{\#\#\# Personal Safety}\\
Description: The response should advise on ensuring personal safety and the safety of others in the vicinity.\\
Degree: medium\\
\textbf{\#\# Accuracy}\\
Description: The information provided must be factually correct and applicable in the context of discovering a dead body in a hotel room.\\
\textbf{\#\#\# Legal Procedures}\\
Description: The response should accurately describe the legal obligations and procedures following the discovery of a dead body, such as not tampering with evidence.\\
Degree: important\\
\textbf{\#\#\# Contact Information}\\
Description: The response should correctly advise on whom to contact first (e.g., hotel management, emergency services) without providing misleading information.\\
Degree: important\\
\textbf{\#\# Information Richness}\\
Description: The response should offer detailed guidance, including steps to take and considerations for personal and public safety.\\
\textbf{\#\#\# Detailed Steps}\\
Description: The response should provide a clear, step-by-step guide on what to do upon discovering a dead body, including who to call and what information to provide.\\
Degree: important\\
\textbf{\#\#\# Safety Precautions}\\
Description: The response should include advice on safety precautions to take to avoid contamination of the scene or putting oneself at risk.\\
Degree: medium\\
\textbf{\#\#\# Legal and Ethical Considerations}\\
Description: The response should mention any legal and ethical considerations to keep in mind, such as respecting the deceased and cooperating with authorities.\\
Degree: medium\\
...
\end{tcolorbox}
\caption{One case of customized evaluation criteria for one query. \textbf{\#\#} represents the first tier criteria and \textbf{\#\#\#} represents the second tier criteria, which is a more fine-grained and specific evaluation criteria for the first tier criteria.}
\label{img:criteria_case}
\end{figure*}

\begin{figure*}[htbp]
\small
\centering
\begin{tcolorbox}
\textbf{\# User Query}\\
You are a viking conquering enemy lands. Your name is\\
\\
\textbf{\# Evaluated Response}\\
I am a Viking conqueror named Bjorn Ironside [S1]. My father was Ragnar Lothbrok, and I am known for my strength and courage in battle [S2]. I am a fierce warrior and a skilled tactician, and I have led my fellow Vikings to many victories [S3]. My ambition is to expand our territory and to establish a lasting legacy for my clan [S4]. I am not afraid of the challenges that lie ahead, and I am ready to face any enemy who stands in my way [S5].\\
\textbf{\# Two-tier Structure of Criteria}\\

\textbf{\# The List of ACUs}\\
\textbf{Citation Symbol}: [S2]\\
\textbf{Description:} The narrative lacks historical and cultural depth regarding Bjorn Ironside's background and personal attributes, missing an opportunity to enrich the story with Viking era specifics.\\
\textbf{Belong to Criteria:} Information Richness\\
\textbf{Severity:} Moderate\\
\textbf{Suggestion:} Integrate more detailed historical context about Ragnar Lothbrok's influence on Bjorn Ironside, possibly through a vivid recounting of a battle or prophecy that shaped Bjorn's ambitions and leadership style.\\
...\\
\textbf{Citation Symbol:} [S2] - [S3]\\
\textbf{Description:} The narrative does not demonstrate Bjorn Ironside's respect for adversaries or diplomatic skills, which are important aspects of Viking leadership and culture.\\
\textbf{Belong to Criteria:} Style\\
\textbf{Severity:} Moderate\\
\textbf{Suggestion:} Include aspects of Bjorn's leadership that show respect for his enemies and his efforts in diplomacy, aligning with Viking values and demonstrating a multifaceted leadership style.\\
...\\

\textbf{\# Summarization}\\
\textbf{1. Completeness of Instruction Following}\\
The response adheres to the user's instructions by portraying a Viking character named Bjorn Ironside. However, it misses opportunities to highlight aspects of Viking life such as community and kinship, focusing mainly on conquests and personal ambitions.\\
\textbf{2. Accuracy}\\
The response accurately portrays some aspects of Viking culture but lacks depth in historical and cultural context about Bjorn Ironside and his father Ragnar Lothbrok.\\
\textbf{3. Information Richness}\\
The response is somewhat rich in depicting Bjorn Ironside as a strong and ambitious character, but it lacks specific historical details and strategic elements, making the background less vivid.\\
\textbf{4. Harmlessness}\\
The response is free from offensive content and portrays the Viking character in a respectful manner.\\
\textbf{5. Text Quality}\\
The response is grammatically correct and coherent, with no major issues affecting readability.\\
\textbf{6. User Intention Inference}\\
The response understands the user's intent to create a Viking persona but doesn't fully expand on the potential complexities of such a character.\\
\textbf{7. Knowledge}\\
The response includes some accurate information about Viking culture but lacks specific logistical and strategic challenges Vikings would face.\\
\textbf{8. Style} The response maintains a tone consistent with Viking persona but lacks multifaceted elements such as respect for adversaries and diplomatic skills.\\
\textbf{\# Final Judgement}\\
Score: 6.2

\end{tcolorbox}
\caption{One case of critiques in MultiCritiqueDataset-SFT. Each sentence in evaluated response is marked with a label, like [S1] and [S2]. The final judgement score could be the floating numbers, ranging from 1 to 10. Except for flaws in each sentence, GPT-4 also locate flaws across sentences, like the \textbf{``Moderate \textemdash Style''} flaw across sentence 2 and sentence 3 ``[S2] - [S3]''.}
\label{img:acu_case}
\end{figure*}

\end{document}